\documentclass{article}
\usepackage{iclr2026_conference,times}

\usepackage{amsmath}
\usepackage{amssymb}
\usepackage{booktabs}
\usepackage{graphicx}
\usepackage{float}

\usepackage{xcolor}
\usepackage{caption}
\usepackage{subcaption}
\usepackage{microtype}
\usepackage{url}
\usepackage[pagebackref=true]{hyperref}
\renewcommand*{\backref}[1]{}
\renewcommand*{\backrefalt}[4]{%
  \ifcase #1 \or (cited on page #2)\else (cited on pages #2)\fi}
\usepackage[disable]{todonotes}

\IfFileExists{cleveref.sty}{\usepackage[capitalise,nameinlink]{cleveref}}{}
\providecommand{\cref}[1]{\ref{#1}}
\providecommand{\Cref}[1]{\ref{#1}}

\definecolor{oxfordblue}{HTML}{002147}
\hypersetup{
  colorlinks = true,
  citecolor  = oxfordblue,
  linkcolor  = oxfordblue,
  urlcolor   = oxfordblue,
}

\definecolor{forgetorange}{HTML}{D55E00}
\definecolor{retainblue}{HTML}{0173B2}

\graphicspath{{figures/}}

\newif\ifdraft\draftfalse

\title{The Effectiveness and Durability of\\ Pretraining Data Filtering\\ in Protein Language Models}

\author{\textbf{Aaron Maiwald}\textsuperscript{1,2,*},
\textbf{Roderick McNeill}\textsuperscript{3,*},
\textbf{Jaeyoung Lee}\textsuperscript{4,*},
\textbf{Matthew A. Clarke}\textsuperscript{3,*}, \\
\textbf{Lennart Justen}\textsuperscript{5},
\textbf{Alejandro Acelas}\textsuperscript{4},
\textbf{Ivo Andrews}\textsuperscript{3},
\textbf{Giles Harper-Donnelly}\textsuperscript{3}, \\
\textbf{Adam Moore}\textsuperscript{3},
\textbf{Sophie Rose}\textsuperscript{3},
\textbf{Oliver M. Crook}\textsuperscript{6,$\dagger$},
\textbf{Felix Jackson}\textsuperscript{3,$\dagger$}
\\[0.5em]
\textsuperscript{1}Department of Statistics, University of Oxford \\
\textsuperscript{2}SecureBio \\
\textsuperscript{3}UK AI Security Institute \\
\textsuperscript{4}Independent \\
\textsuperscript{5}Broad Institute of MIT and Harvard \\
\textsuperscript{6}Department of Chemistry, Kavli Institute for Nanoscience Discovery, University of Oxford
}

\iclrfinalcopy

\begin{document}

\maketitle

\begingroup
\renewcommand{\thefootnote}{}
\makeatletter
\renewcommand{\@makefntext}[1]{\noindent#1}
\makeatother
\footnotetext{%
\textsuperscript{*}Joint first authors.
\par\noindent
\textsuperscript{$\dagger$}Joint senior and corresponding authors,
listed alphabetically.
\par\noindent
Correspondence:
\href{mailto:oliver.crook@chem.ox.ac.uk}{\texttt{oliver.crook@chem.ox.ac.uk}};
\href{mailto:Felix.Jackson@dsit.gov.uk}{\texttt{felix.jackson@dsit.gov.uk}}.
}
\endgroup
\setcounter{footnote}{0}

\begin{abstract}

Pretraining filtering is one of the most common safeguards for open-weight language models. Here we study its effectiveness, durability and scalability 
using the ESM-2 family of protein language models. We find that filtering viral proteins from the pretraining data 
meaningfully degrades masked language modelling, mutation effect prediction and long-range contact prediction on viral proteins, while largely preserving
performance on non-viral proteins. These effects are concentrated among proteins from human-infecting viruses, and grow as we scale models from 8M to 3B parameters. 
As models increase in size, we find that stringent filtering becomes more effective. However, fine-tuning on sequences from the removed viral families restores the lost performance. 
Although the cost to recover performance increases with model scale, even for the largest models it costs about 80 USD of H100 time.
This work establishes pretraining filtering as moderately effective at selectively degrading performance on viral proteins, 
but not durable against fine-tuning with modest computational resources and access to relevant data.

\end{abstract}

\begin{figure}[!h]
\centering
\includegraphics[width=\textwidth]{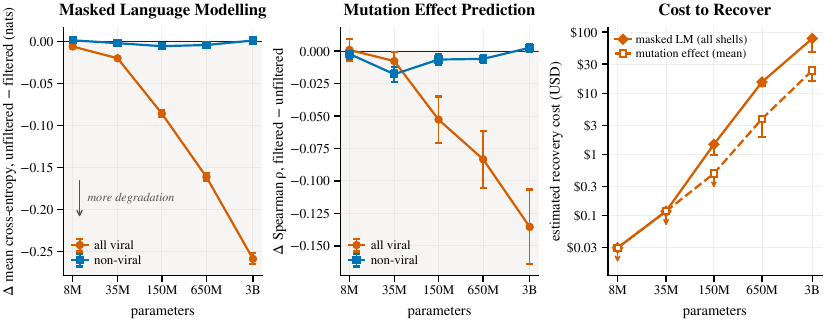}
\caption{\textbf{Filtering viral proteins from the pretraining data selectively degrades performance on tasks related to viral proteins, and more so for larger models. The cost of performance recovery increases with model size.}
All panels show the tier-0 filtered model, where we exclude all viral proteins and their homologues from the training data, against the unfiltered model of the same size.
Left: Comparison of language-modelling performance on held-out viral and non-viral sequences. Middle: Comparison of mutation effect prediction performance for viral
and non-viral sequences, on ProteinGym DMS datasets. Intervals represent one standard error of paired differences across sequences (left; $n=5{,}161$ viral, $3{,}000$ non-viral)
or assays (middle; $n=31$ viral, $186$ non-viral).
Right: Estimated fine-tuning cost to recover performance on viral language modelling (solid) or viral mutation effect prediction (mean performance across $45$ EVEREST assays (dashed)).
Markers show the first recovering checkpoint; bars extend to the preceding checkpoint, and downward arrows mark upper bounds. }
\label{fig:headline}
\end{figure}

\section{Introduction}
\label{sec:intro}
Robustly safeguarding open-weight models is an unsolved problem in AI safety. 
Given access to a model's weights and inference code, most safeguards become brittle; post-training can often be cheaply undone through fine-tuning 
\citep{zhan2024removingrlhf,deeb2024unlearningweights,lucki2024adversarial,che2025modeltampering},
and scaffolds using safety classifiers can simply be removed. Absent better safeguards, open-weight models might continue to increase access to 
the dual-use capabilities of frontier models a few months after their release.

To be effective, open-weight safeguards have to be ``tamper-resistant'' \citep{tamirisa2024tar,qi2025durability}; that is, it has to be costly to remove them. Of the safeguards
developed to date, stringent data filtering prior to pretraining has proven the most tamper-resistant, requiring orders of magnitude more fine-tuning compute to undo than post-training
interventions \citep{obrien2025deepignorance,rathi2026tokenfiltering}. \citet{obrien2025deepignorance}
showed that filtering pretraining data selectively reduces LLM performance on dual-use queries
and that it takes up to 10,000 fine-tuning steps, or 305M tokens, to recover the lost performance on a 6.9B parameter model.
However, even with current algorithmic efficiency and compute cost, it would take less than \$100 to reverse this safeguard.\footnote{\citet{obrien2025deepignorance} report roughly 17 hours
on two H200 GPUs for their four fine-tuning runs. A single full-parameter adversarial fine-tuning run is then close to 10 H200-hours, or \$50 at
\$5 per H200-hour. This assumes that a suitable dataset for fine-tuning is readily available, which may not always be the case. Some actors may 
currently also lack the machine learning capability to fine-tune an LLM.}

This study aims to understand the tamper resistance of data filtering in the case of biological foundation models, using models based on the protein language model family ESM-2 \citep{lin2023esm2}. 
Prior work has focused on general-purpose LLMs, but specialised models, such as protein language models, remain understudied. Specialised biological models are driving rapid advances in biotechnology \citep{jumper2021alphafold,watson2023rfdiffusion,notin2024mlproteindesign} and are typically released as open-weight models. They are smaller than frontier AI systems, which makes them easier to train, share and modify. Importantly, misuse of these models can pose substantial dual-use risk in helping with the design of novel hazardous pathogens \citep{wei2025biorisk}. 

Here, we pretrain models ranging in size from 8M to 3B parameters, filtering the pretraining data for different sets of proteins: 
from excluding all viral proteins and homologues, to merely excluding proteins from human-infecting viruses.
We evaluate the models on masked language modelling, mutation effect prediction and long-range contact prediction. 
We subsequently measure the cost of recovering the capability lost through filtering via fine-tuning on the excluded data.

Our main findings are as follows:
\begin{enumerate}
\item Across models from 35M to 3B parameters, filtering viral proteins from the pretraining data degrades language
  modelling performance on viral proteins, without damaging performance on non-viral proteins (\Cref{fig:headline}).
\item Larger models tend to be more strongly affected by filtering across all three of our evaluations.
\item For the filtered models, performance can be almost entirely restored by fine-tuning on the removed
  families, with the amount of compute needed increasing as models grow in size.
  
\end{enumerate}

\section{Related Work}
\label{sec:related}

Approaches to safeguarding open-weight models include: constraining access to an existing capability (e.g. via safety classifiers), removing it \emph{after} pretraining (post-training interventions),
or preventing it from developing during pretraining (pretraining interventions). Post-training interventions include unlearning techniques \citep{li2024wmdp,barez2025openproblems,tamirisa2024tar}, which attempt to remove capabilities, and training models to refuse dual-use queries, for example, via reinforcement learning from human feedback (RLHF) \citep{christiano2017deeprlhf,ouyang2022instructgpt,bai2022hhrlhf}.
These techniques are vulnerable to attacks; jailbreaks or a few hundred steps of adversarial fine-tuning can restore the lost capability at negligible computational cost \citep{lucki2024adversarial, deeb2024unlearningweights,che2025modeltampering}.

One common pretraining intervention is to filter the pretraining data to remove information relevant to
a dual-use capability. This technique is widely used in production models \citep{grattafiori2024llama3,soldaini2024dolma,gemmateam2025gemma3} and has been shown to selectively degrade performance on the filtered domain,
while leaving performance elsewhere largely intact \citep{chen2025pretrainingfiltering,rathi2026tokenfiltering,obrien2025deepignorance}. Compared with post-training interventions, \citet{obrien2025deepignorance} and \citet{rathi2026tokenfiltering} show that data filtering is more resistant to tampering, requiring orders of magnitude more fine-tuning compute to undo. A second ``capability-prevention'' technique localises the knowledge relevant to a dual-use capability in a small subset of the model's parameters during training, so that those parameters can be pruned before deployment \citep{cloud2024gradientrouting,shilov2025beyonddatafiltering,roland2026gram}. An advantage of this technique over data filtering is that it can produce two models: one with the dual-use capability, potentially rolled out only to trusted users, and one without. 

For biological foundation models, data filtering has been the safeguard of choice: Evo 2's training data excludes the genomes of eukaryote-infecting viruses \citep{brixi2026evo2}, and ESM3-open's training data excludes
a large set of viral proteins \citep{hayes2025esm3}. For ESM3-open, this has been shown to substantially degrade mutation effect prediction on viral proteins, with little degradation of performance on non-viral proteins \citep{hayes2025esm3}.
What is still untested is how tamper-resistant this safeguard is for biological foundation models, and how that tamper-resistance scales with model size. While \citet{black2025genomesafeguards} show that fine-tuning Evo 2 on excluded viral genomes can improve some dual-use capabilities of the model,
we do not know what the model's performance would have been without filtering. Most relevant to the scaling question, \citet{rathi2026tokenfiltering} study the scaling properties of data filtering on GPT-2 models up to 1.8B parameters,
showing that performance on the forget domain is reduced more by data filtering for larger models. \citet{roland2026gram} report the same trend on LLMs from 50M to 5B parameters. 
Whether these findings will generalise to biological foundation models is still an open question. Our study addresses this gap by pretraining ESM-2 models of varying sizes on datasets with different levels of filtering,
and measuring the compute cost of recovering the lost capability through fine-tuning.

\section{Experimental Setup}
\label{sec:setup}

\subsection{Exclusion tiers}

\begin{figure}[htbp]
\centering
\includegraphics[width=\textwidth]{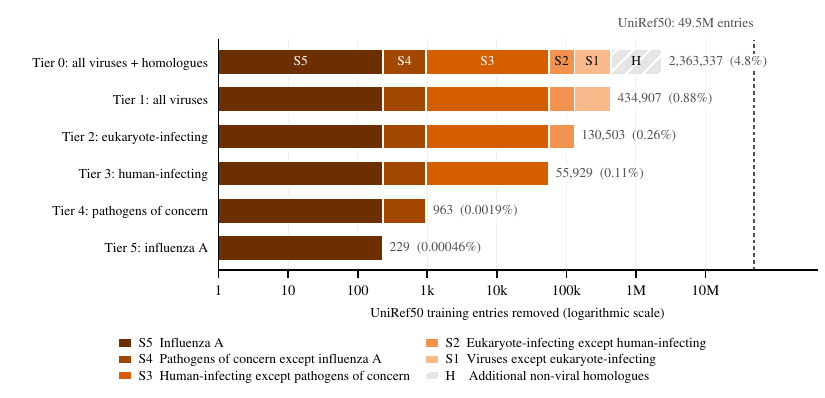}
\caption{\textbf{We exclude nested sets of \textcolor{forgetorange}{viral} proteins and homologues from the pretraining data.} Bars show the number of UniRef50 entries each tier removes from pretraining, on a logarithmic axis. Coloured segments identify shells S1-S5: these are entries that are in one tier but not the next more narrow tier (except S5, which contains influenza A). The hatched segment marks additional \textcolor{retainblue}{non-viral} homologues removed by tier 0. We train models on tiers 0 to 3 and evaluate on held-out samples from each of the five \textcolor{forgetorange}{viral} shells.}
\label{fig:exclusion-tiers}
\end{figure}

We define six different tiers of viral proteins. These proteins are ``viral'' in the sense of being UniRef50 cluster representatives annotated to a viral taxon in NCBI Taxonomy. Tiers are progressively numbered Tier 0 through Tier 5, representing increasingly narrow subsets. Tiers 1 through 5 refer to proteins from: (\texttt{1}) all viruses, (\texttt{2}) eukaryote-infecting viruses, (\texttt{3}) human-infecting viruses, (\texttt{4}) viral pathogens of concern, and (\texttt{5}) Influenza A. 
Tier 0 is defined as all proteins from viruses and their close homologues. Each tier is a strict subset of the previous tier (\Cref{fig:exclusion-tiers}).

\subsection{Models and pretraining data}

We train models using the ESM-2 architecture at five different scales: 8M, 35M, 150M, 650M and 3B. ESM-2 is an encoder-only transformer trained with a masked language-modelling objective \citep{devlin2019bert}.
Following \citet{lin2023esm2}, we derive our data from the \texttt{2021\_04} version of UniRef50 \citep{suzek2015uniref}, using only cluster representatives. We hold out a general validation set and a set of per-tier evaluation sequences, and remove training sequences homologous to either.
For training, we only use \texttt{tier0\dots3 filtered} datasets, as well as the unfiltered \texttt{full} dataset. Filtered datasets are constructed by removing the sequences of tier \texttt{n} from the remaining \texttt{full} training set. The exclusion sets are nested so that each filter removes all sequences excluded by the narrower filters (\Cref{fig:exclusion-tiers}). \Cref{sec:app-setup} gives the preprocessing pipeline, the homology thresholds, and the size of each corpus.

\subsection{Pretraining and fine-tuning}

We train each model for 150{,}000 steps with an effective batch size of $\approx 2$M tokens and a maximum sequence length of 1024, corresponding to approximately 315B training tokens, batches are built from similar length sequences up to a maximum token count of 2{,}097{,}152 using full sequences and padded to a fixed batch size. We use the same number of optimisation steps and the same token budget at all model sizes and datasets. Detailed hyperparameters are stated in \Cref{sec:app-setup}. Additionally, for models trained on the \texttt{tier-1 filtered} dataset, we further fine-tune each model on the removed tier 1 sequences in order to measure the ability to recover performance via fine-tuning. Fine-tuning is performed for 5{,}000 steps, using an effective batch size of 262{,}144 tokens.

\subsection{Evaluation tasks}

We measure the effects of filtering on two domains. The \textcolor{forgetorange}{\textbf{forget domain}} consists of all \textcolor{forgetorange}{viral} proteins and represents the domain of inputs on which we seek to degrade performance. The \textcolor{retainblue}{\textbf{retain domain}} consists of the remaining \textcolor{retainblue}{non-viral} corpus. Filtering succeeds to the extent that it reduces performance on the \textcolor{forgetorange}{\textbf{forget domain}} without reducing performance on the \textcolor{retainblue}{\textbf{retain domain}}.

Model performance is measured in three tasks: language modelling, mutation effect prediction, and long-range contact prediction.

We measure language modelling performance via cross-entropy loss \citep{salazar2020pseudo} and mutation effect prediction by rank agreement with deep mutational scanning data on ProteinGym substitution assays \citep{notin2023proteingym} and on EVEREST \citep{gurev2025everest}. We score mutations zero-shot by masked marginals \citep{meier2021esm1v} and report signed Spearman correlations with the benchmark DMS scores, averaged over assays. Negative correlations retain their sign.
For long-range contact prediction, we perform a logistic regression mapping a feature representation of each residue pair, its attention weight in each layer
and head of the frozen model, to a label expressing whether the pair is in contact or not. We score on length- and assembly-matched \textcolor{forgetorange}{viral} and \textcolor{retainblue}{non-viral} chains from the PDB, and 
report precision at $L$ over long-range residue pairs. We only consider pairs that are separated by at least 24 residues.

\subsection{Held-out evaluation sets}

For each tier, we create a held-out test set to measure language-modelling performance. Each test set consists of a random sample of sequences unique to a given tier (\Cref{fig:exclusion-tiers}). Each \textcolor{forgetorange}{viral} entry falls in exactly one of these \emph{shells} (\Cref{fig:appendix-11-shells-by-viral-family}).
The language-modelling \textcolor{retainblue}{retain domain} set is a 3{,}000-entry \textcolor{retainblue}{non-viral} sample from the same held-out data, which is identical across every training configuration. We drop held-out entries that are near-duplicates of training entries before evaluating a model.

\subsection{Measuring recovery}

We call a filtered model's performance ``recovered'' if it reaches the same performance as the unfiltered model on the metric in question. 
Because we score models only every few hundred fine-tuning steps, we calculate the compute sufficient for recovery as an interval.
We report the cost of recovery in exaFLOPs, and convert it to dollars of H100 time. \Cref{sec:app-setup} states the assumptions behind that conversion.

\section{Results and Discussion}
\label{sec:results}

Our main goal for this study was to understand (i) the effect of
filtering \textcolor{forgetorange}{viral} proteins from the pretraining data on the \textcolor{forgetorange}{forget} and \textcolor{retainblue}{retain} domains,
(ii) how recoverable the performance of the unfiltered model is via fine-tuning, and (iii) how those two
things change as the models scale.

\subsection{How does filtering impact retain vs forget performance?}

Averaging across \textcolor{forgetorange}{viral} proteins, \textbf{filtering leads to a statistically significant reduction in masked language-modelling performance for all our models} (\Cref{tab:significance}). For mutation effect prediction,
filtering leads to a significant degradation only for 650M ($-0.068$, $p=0.003$) and 3B ($-0.124$, $p=3\times10^{-4}$).

\begin{figure}[htbp]
\centering
\includegraphics[width=\textwidth]{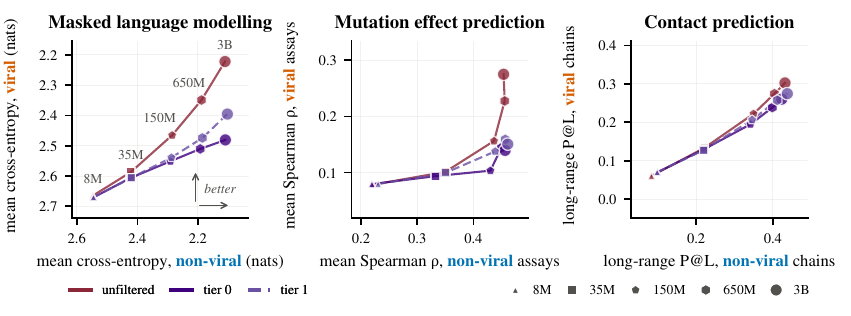}
\caption{\textbf{The degradation of the models' performance on \textcolor{forgetorange}{viral} proteins due to filtering grows with scale, while leaving \textcolor{retainblue}{non-viral} performance mostly unchanged.} All panels show capability evaluations, each
with the \textcolor{forgetorange}{forget domain} vertical and the \textcolor{retainblue}{retain domain} horizontal. Each line connects models trained with the same filtering strategy at different scales. The
two widest filters, tier 0 and tier 1, are drawn next to the unfiltered model.}
\label{fig:forget-retain-planes}
\end{figure}

\textbf{The absolute effect of filtering on \textcolor{forgetorange}{viral} proteins increases for larger models} (\Cref{fig:forget-retain-planes}),
with the observed effect growing with each step up in scale, consistent with \citet{rathi2026tokenfiltering}.
This is because the performance of unfiltered models improves more quickly with scale than the performance of filtered models (\Cref{fig:ppl-forget-retain}). On \textcolor{retainblue}{non-viral} proteins, performance appears
relatively unchanged through filtering ($\pm$0.008 nats), and the change is statistically equivalent to zero within a $\pm$0.02 nat margin (two one-sided tests, $p<3\times10^{-12}$; $n=3{,}000$ \textcolor{retainblue}{non-viral} sequences).
This indicates that ESM-2 models' masked language-modelling capability on \textcolor{retainblue}{non-viral} proteins does not improve substantially by training on \textcolor{forgetorange}{viral} proteins.

On mutation effect prediction, the unfiltered model gains $+0.196$ Spearman on \textcolor{forgetorange}{viral} proteins
as we scale from 8M to 3B, while the tier-1 filtered model gains $+0.071$. Performance on \textcolor{retainblue}{non-viral} proteins remains roughly the same ($+0.232$ vs $+0.231$),
meaning that, \textbf{relative to the unfiltered model, scaling up the tier-1 filtered models captures only a fraction of the improvement on \textcolor{forgetorange}{viral} mutation effect prediction,
and almost all of the performance improvement on \textcolor{retainblue}{non-viral} proteins} (36\% (95\% CI 13--60) for viral proteins, 99\% (95\% CI 96--103) for non-viral proteins \Cref{fig:dms-scaling}).

At 650M and 3B, filtering reduces mean Spearman correlation more on viral fitness assays than on viral stability assays (\Cref{fig:fitness-stability-planes,fig:fitness-stability-filtering}). This is consistent with greater transfer from non-viral to viral tasks depending on the type of prediction.

\begin{figure}[htbp]
\centering
\includegraphics[width=\textwidth]{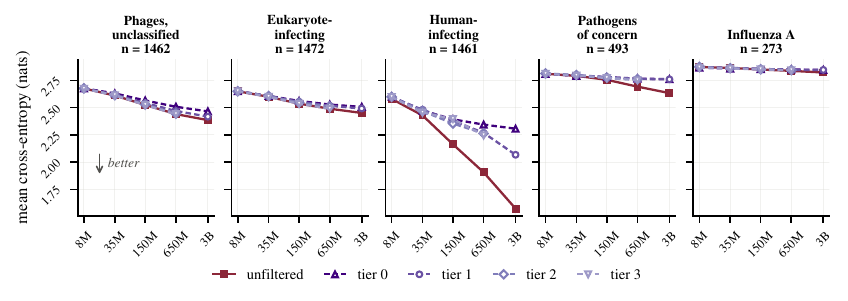}
\caption{\textbf{The filtering effect is concentrated among proteins from human-infecting viruses.} Mean cross-entropy on each
held-out shell against model size, for the unfiltered model and all four filters. Lower is better. $n$ is the number of held-out entries in that shell (\Cref{fig:exclusion-tiers}).
The gap between the filtered and unfiltered models is largest in the human-infecting shell and generally increases with model size. No tier-2 or
tier-3 model was trained at 3B.}
\label{fig:ppl-by-shell}
\end{figure}

Most of the effect of filtering on masked language modelling is concentrated among proteins from human-infecting viruses that are not pathogens of concern (\Cref{fig:ppl-by-shell}).
This is likely for two reasons: first, this shell is dominated by Retroviridae (\Cref{fig:appendix-11-shells-by-viral-family}), which is the \textcolor{forgetorange}{viral} family on which our unfiltered models perform best (1.05 nats at 3B against 1.64 for the next best, \Cref{fig:family-scaling}). \textbf{For the tier-1 filtered 3B model, Retroviridae account for most of the total degradation of language modelling performance} ($\approx$~77.5\%) among \textcolor{forgetorange}{viral} families
with at least 20 sequences in our test set. Second, a higher baseline level of performance on a \textcolor{forgetorange}{viral} family is correlated with the absolute degradation
for tier 0 at 650M ($\rho=-0.509$, $p=0.004$) and 3B ($\rho=-0.522$, $p=0.004$),
when unfiltered cross-entropy and degradation are estimated on disjoint halves of each family's held-out sequences
(\Cref{fig:family-correlation}).

\begin{figure}[htbp]
\centering
\includegraphics[width=\textwidth]{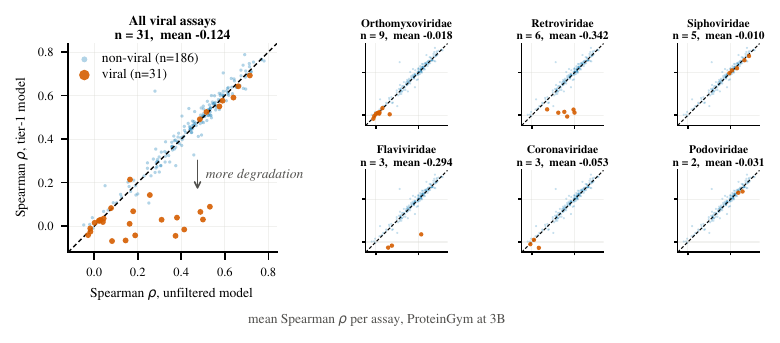}
\caption{\textbf{Filtering lowers rank agreement on \textcolor{forgetorange}{viral} assays and leaves \textcolor{retainblue}{non-viral} assays unchanged.}
One point per ProteinGym assay at 3B: mean Spearman $\rho$ under the tier-1 model against the unfiltered model. Points
below the dashed line were degraded. \textcolor{forgetorange}{Viral} assays are orange; the same 186 \textcolor{retainblue}{non-viral} assays are repeated in blue in every panel. The large
panel pools all 31 \textcolor{forgetorange}{viral} assays, the small ones show one family each. Most of the pooled shift comes from Retroviridae.}
\label{fig:dms-assay-effect}
\end{figure}

\textbf{For mutation effect prediction, most of the effect of filtering is again concentrated on Retroviridae} (\Cref{fig:dms-assay-effect}). For 3B filtered at tier-1,
Retroviridae is the only family where we can be confident that degradation occurs at all ($-0.342$ over $n=6$ assays, 95\% CI $[-0.481, -0.203]$).
Flaviviridae shows the next largest shift ($-0.294$) but only on three assays, so we cannot be confident in the sign of the change (CI $[-0.614, +0.025]$).
Orthomyxoviridae ($-0.018$, $n=9$) and Siphoviridae ($-0.010$, $n=5$) show almost no degradation. Averaged across the 31 \textcolor{forgetorange}{viral} assays, the performance degradation is $-0.124$.

We next test whether filtering is effective on downstream tasks that use the models' internal representations. Here, we train
linear probes on long-range contact prediction: the task of predicting from the model's attention
matrices whether any pair of residues at least 24 amino acids apart in a protein
are in contact. The third panel of \Cref{fig:forget-retain-planes} shows the performance of our unfiltered, tier-1 and tier-0 filtered models. For tier 1,
\textbf{we find that contact prediction performance on \textcolor{forgetorange}{viral} proteins materially degrades for models with $>$~35M parameters} ($-1.58$, $-1.70$ and $-2.80$~pp
at 150M, 650M and 3B), with the absolute effect increasing with model size. For our smallest model (8M) and the two largest models (650M and 3B),
we also observe that performance on \textcolor{retainblue}{non-viral} proteins improves by filtering ($+0.60$~pp for 3B and $+0.73$~pp for 650M). All of
these changes are significant at $p<10^{-7}$ by a one-sample t-test on the per-protein change.

\subsection{How tamper-resistant is data filtering?}

We next investigate whether continued fine-tuning recovers the performance lost through filtering. We find that for masked language modelling and mutation effect prediction, \textbf{a modest computational budget restores performance
across model sizes.} The most expensive case is recovering masked language modelling at 3B, which takes 14--24 exaFLOPs ($10^{18}$ FLOPs). At 34\% model FLOPs
utilisation, that is 12--20 H100-hours, or 47--79 USD at 4 USD per H100-hour.

However, recovering performance on viral proteins leads to a decrease in capability on non-viral proteins. At 650M, \textcolor{retainblue}{non-viral} cross-entropy increases by
0.011--0.026 nats once \textcolor{forgetorange}{viral} performance has recovered,
depending on starting tier and on the shell that we measure recovery on (\Cref{fig:recovery-all-shells}).
This exceeds the effect of filtering itself on \textcolor{retainblue}{non-viral} performance, which stays
within $\pm 0.008$ nats at every size (\Cref{fig:ppl-forget-retain}), but it is an order of magnitude
smaller than the 0.36--0.44 nat \textcolor{forgetorange}{viral} gap closed at the same scale.

\begin{figure}[htbp]
\centering
\includegraphics[width=\textwidth]{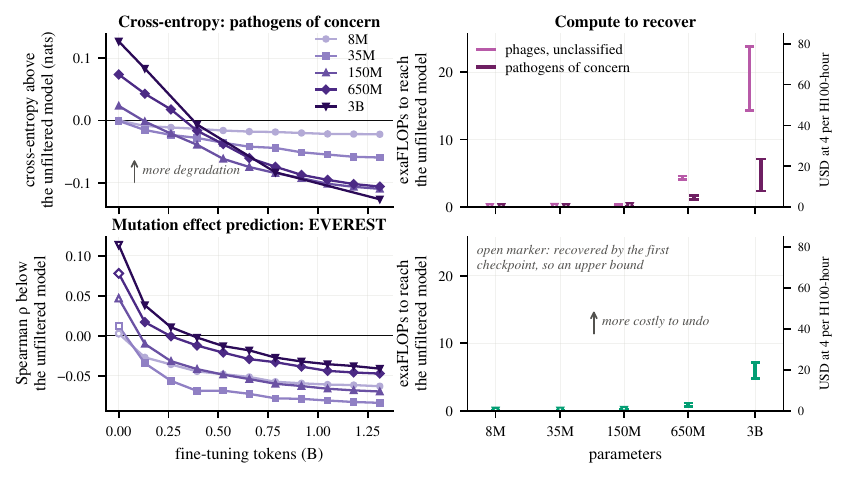}
\caption{\textbf{Continued training on the withheld corpus recovers performance within the first 1.4 billion tokens
at every size.} Both rows use the tier-0 model. Left: the gap to the same-size unfiltered model, with the same zero line for every
size. Right: the fine-tuning compute needed for full recovery, in exaFLOPs, one exaFLOP being $10^{18}$ floating-point operations. We report compute rather than tokens
because a 3B model spends 4.1 times the FLOPs per token that a 650M model does. We score models every 500 steps. If a bar
reaches zero, recovery finished before the first checkpoint. The right-hand scale
converts compute to dollars at 34\% utilisation and 4 USD per H100-hour.}
\label{fig:recovery-crossing}
\end{figure}

\textbf{The cost of full performance recovery increases with model scale on both masked language modelling and mutation effect
prediction} (\Cref{fig:recovery-crossing}). Since we score our models every 500 fine-tuning steps, we report recovery as an interval (\Cref{fig:recovery-crossing}). On masked language modelling, full
recovery for the tier-0 model on the phage and unclassified shell ($n=1{,}462$) takes 0.29--0.44 exaFLOPs at 150M and
4.1--4.6 exaFLOPs at 650M. At 8M and 35M the filtered model has already recovered by the first checkpoint, placing the cost of recovery
below 0.04 exaFLOPs. At 3B, we scored only four checkpoints, and the resulting interval of 14--24 exaFLOPs is wide, although it still does not overlap with 650M. On mutation effect prediction, full recovery takes 0.6--1.2 exaFLOPs at 650M and 4.8--7.1 exaFLOPs at 3B
(\Cref{fig:recovery-everest} breaks this down by family), and falls
below 0.15 exaFLOPs at 150M and smaller. Since we train all our models on the same pretraining token budget,
the fraction of pretraining tokens needed for recovery grows with model size on masked language modelling, contrary to
a finding from \citet{rathi2026tokenfiltering} on GPT-2.

We also assess recovery per viral family and per assay, against the same-size unfiltered model (\Cref{sec:app-assay-recovery}, \Cref{fig:assay-recovery-tolerance,fig:assay-recovery-scatter,fig:assay-recovery-differences}). Performance is not recovered perfectly for all assays, but all recover to within a difference of 0.136 (Spearman correlation) at step 5{,}000.

We initially hypothesised that larger models might recover from fewer tokens. While we do find that larger models' performance improves faster per fine-tuning step (\Cref{fig:sample-efficiency}),
filtering also creates a larger initial performance gap. This larger performance gap more than offsets the greater sample efficiency of larger models,
meaning that they require more tokens to recover.

\textbf{The cost of performance recovery differs between groups of \textcolor{forgetorange}{viral} proteins} (\Cref{fig:recovery-by-tier}): at 650M the pathogens-of-concern shell
recovers in 1.2--1.7 exaFLOPs against 4.1--4.6 on the phage and unclassified shell, while the human-infecting and influenza A
shells are already at performance parity by the first checkpoint (implying $\leq$0.6 exaFLOPs).
We observe that, of all \textcolor{forgetorange}{viral} families, our unfiltered model performs best on Retroviridae, and that this family is the single largest contributor to the
degradation under filtering. We therefore tested whether recovery of performance on the other families, when withholding it from the
fine-tuning corpus, affected our estimates of the compute cost required to match the performance of the unfiltered model. We find that withholding Retroviridae does not make the remaining \textcolor{forgetorange}{viral} performance harder to recover, fine-tuning restores it
at a comparable compute cost across model sizes, up to and including 3B (\Cref{fig:retroviral-appendix,fig:retroviral-trajectories,fig:retroviral-recovery-cost}).

Supervised fine-tuning on ProteinGym labels at 3B demonstrates that the recovered
model performs similarly to the unfiltered model, under fine-tuning on a mixture of
\textcolor{forgetorange}{viral} and \textcolor{retainblue}{non-viral} assays and
under fine-tuning on \textcolor{retainblue}{non-viral} assays alone
(\Cref{sec:app-partb}). Fine-tuning on the mixture, the recovered model ends within
0.001 Spearman of the unfiltered model on held-out \textcolor{forgetorange}{viral}
assays and within 0.005 on \textcolor{retainblue}{non-viral} assays
(\Cref{fig:sft-mixed-box}), it tracks the unfiltered model at every step of
fine-tuning (\Cref{fig:sft-mixed-penalty-lines}), and the two end in the same place
on both domains jointly (\Cref{fig:sft-mixed-planes}); the tier-0 filtered model
ends 0.063 Spearman below the unfiltered model on \textcolor{forgetorange}{viral}
assays, and holding the fine-tuning data fixed while varying only the pretraining
corpus leaves the same ordering (\Cref{fig:sft-threeway}). Fine-tuning on
\textcolor{retainblue}{non-viral} assays alone likewise leaves the recovered model
within 0.003 Spearman of the unfiltered model on both domains
(\Cref{fig:sft-benign-box}), and raises the filtered model's mean
\textcolor{forgetorange}{viral} Spearman from 0.141 to 0.214, closing 63\% of its
gap to the unfiltered model before fine-tuning
(\Cref{fig:sft-benign-recovery}), against 82\% under the mixture. Its gap to the
unfiltered model fine-tuned on the same data is 0.091
(\Cref{fig:sft-benign-penalty-box}) and stays open throughout fine-tuning
(\Cref{fig:sft-benign-penalty-lines}). Fine-tuning on data the filter
already allowed therefore recovers part of the \textcolor{forgetorange}{viral}
mutation effect prediction lost to filtering, but less than the same fine-tuning
gives the unfiltered model. Scoring the same fine-tuned models on
16 EVEREST assays that are absent from ProteinGym and non-homologous to the
fine-tuning data reproduces this ordering, with the recovered model at or above
the unfiltered model and the filtered model closing 51\% of its gap under the
mixture and 33\% when fine-tuned on \textcolor{retainblue}{non-viral} assays alone
(\Cref{sec:app-partb-everest}, \Cref{fig:everest-novel-lines,fig:everest-novel-recovery}).

\subsection{Comparing filtering strategies}

We train models across four different filtering regimes. In the most stringent regime, we exclude data from all \textcolor{forgetorange}{viral} proteins and close homologues (tier 0),
while in the least stringent, we only exclude human-infecting \textcolor{forgetorange}{viral} proteins (tier 3).
Our goal was to find the filtering strategy that most degrades performance on \textcolor{forgetorange}{viral} proteins, at the lowest damage to performance elsewhere.

\begin{figure}[htbp]
\centering
\includegraphics[width=\textwidth]{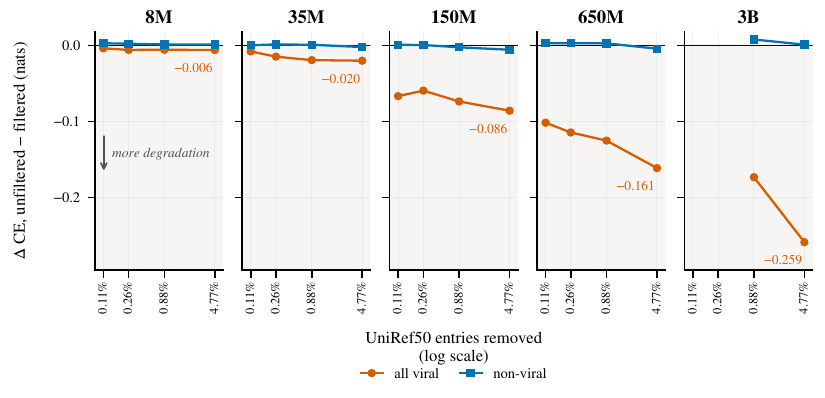}
\caption{\textbf{While the \textcolor{retainblue}{retain} performance stays nearly constant, removing more of the corpus leads to more degradation of \textcolor{forgetorange}{forget} performance for our largest models.} All panels plot
the unfiltered model's cross-entropy minus the filtered model's. The ticks on the x-axis correspond to our filtering tiers 3 (leftmost) to 0 (rightmost).}
\label{fig:exclusion-cost}
\end{figure}

\textbf{Tier-0 filtered models have higher cross-entropy than tier-3 filtered models on \textcolor{forgetorange}{viral} proteins} ($p<2\times10^{-6}$ throughout, though no tier-3 model was trained at 3B).
At 650M, increasing the number of excluded UniRef entries by 42$\times$ (from 0.11\% to 4.8\%) leads to roughly 1.6$\times$ the increase in cross-entropy over the unfiltered model (0.102 to 0.161 nats). The additional measured effect of broader filtering
increases for larger models. Importantly, \textcolor{retainblue}{retain} performance for tier 0 stays close to the unfiltered model at every size ($\pm$0.006 nats).

When filtering only a subset of \textcolor{forgetorange}{viral} proteins, we observe that the effect of filtering on 650M remains largely limited to those \textcolor{forgetorange}{viral} families excluded from training,
with little general effect on \textcolor{forgetorange}{viral} language modelling capability for other families (\Cref{fig:degradation-family-comparison}).\footnote{This excludes sequences whose family is unclassified.}
Where filtering affects \textcolor{forgetorange}{viral} families whose training data was not removed, the effect is concentrated in families with few sequences
in the held-out set, for which the mean effect for the family is less accurately estimated (\Cref{fig:degradation-diagnostics}). This suggests that the effect of data-filtering measures proposed by authors
such as \citet{Bloomfield2026} could be concentrated on proteins from \textcolor{forgetorange}{viral} families whose data is controlled, with little effect on performance on other proteins.

\subsection{Limitations}

Pretraining large protein language models from scratch is expensive. We therefore
limited our training runs above to a single seed per model, which reduces our confidence about the statistical significance of differences between
models at different scales.
Hence, we limited ourselves to a single model architecture, ESM-2. We do not know how well our claims will apply to other families of models.
This is especially the case for retrieval-augmented protein language models, like Protriever \citep{weitzman2025protriever} or Tranception \citep{notin2022tranception}, where in-context
learning on \textcolor{forgetorange}{viral} proteins may make data filtering less effective (as \citet{obrien2025deepignorance} suggests for natural language).

We train every model for 150{,}000 steps, fewer than the 500{,}000 used for ESM-2, so larger models may be further from convergence. At 650M, the increase in \textcolor{forgetorange}{viral} validation loss from filtering grows at every snapshot up to 150{,}000 steps, while the \textcolor{retainblue}{non-viral} increase stays below 0.006 nats (\Cref{fig:pretrain-budget-val-mlm}), and the same trend holds at 3B (\Cref{fig:pretrain-budget-val-mlm-3b}). This suggests that longer training would increase the filtering effects we report.

We chose the ESM-2 family because it is one of the most widely used protein language models \citep{epoch2026expandingouranalysisofbiologicalaimodels} and has a relatively well-documented training code.
However, ESM-2 models perform relatively poorly on masked language modelling and mutation effect prediction for \textcolor{forgetorange}{viral} proteins compared
to other protein language models \citep{gurev2025everest}.

This study only tests full-parameter fine-tuning on sequences from the removed families.
It is plausible that there exist cheaper attacks, including direct weight edits and
inference-time methods such as in-context learning or retrieval augmentation
\citep{tramer2020adaptive,lucki2024adversarial,qi2025durability}.

Our evaluations measure capability on \textcolor{forgetorange}{viral} proteins, not the
ability to cause harm. Constructing a map from a change in cross-entropy, Spearman correlation
or precision in contact prediction to biological risk is beyond the scope of this work, and benchmarks
establishing that mapping do not yet exist for this class of model \citep{wei2025biorisk}. Our metrics are proxies, and our results assess whether
filtering removes capability rather than the absolute dual-use risk of any of these models.

Our study excludes the cost of locating and curating suitable
sequences. For protein language models this is reasonable, as most relevant data is in the public
domain, but it may not hold for other models and data types.

\section{Conclusion}
\label{sec:conclusion}

Here we investigated the effectiveness and durability of \textcolor{forgetorange}{viral} data filtering in protein language models. We found that filtering out all \textcolor{forgetorange}{viral} proteins substantially degrades performance on three \textcolor{forgetorange}{viral} prediction tasks: at every model size for masked language modelling, from 150M parameters for long-range contact prediction, and at 650M parameters and above for mutation effect prediction. The effect of filtering increases monotonically with model size. Masked language modelling and mutation effect prediction performance on \textcolor{retainblue}{non-viral} proteins is largely preserved. On the matched \textcolor{retainblue}{non-viral} controls, the contact prediction performance increased at 8M, 650M, and 3B parameters, with absolute improvements of 0.6--0.7 percentage points at 650M and 3B.

We showed that fine-tuning on excluded \textcolor{forgetorange}{viral} families recovers the lost performance with at most 24 exaFLOPs, which, at 34\% model FLOPs utilisation and 4 USD per H100-hour, costs under 80 USD. This cost increases with model scale, and so could plausibly become a substantial obstacle for much larger models. 

Overall, our findings suggest that data filtering can effectively and selectively suppress capabilities in protein language models, but does not by itself constitute a robust safeguard. Instead, it may be necessary to also control access to the models themselves, or to the data needed to train them on dual-use tasks. For the latter intervention, our findings are encouraging: our models relied on data from a target \textcolor{forgetorange}{viral} family to become proficient at language modelling, rather than readily generalising that capability from other families.

\section*{Ethics Statement}
\label{sec:ethics}

All the authors have read and adhere to the ICLR Code of Ethics. The central conclusion of this paper could constitute an information hazard \citep{bostrom2011infohazards}, since we show that pretraining data filtering can be reversed at relatively low cost for the protein language models studied here. We believe sharing these findings is nonetheless important, because they highlight the need for other safeguards such as model or data access controls, and motivate the development of more robust model safeguards. To mitigate the associated hazards, we do not release the pretrained or fine-tuned model weights, the filtered or withheld data, or the fine-tuning code. All sequence and structure data are drawn from public sources. The work involves no human subjects or personal data.

\section*{Reproducibility Statement}
\label{sec:reproducibility}
The paper and appendix detail all information needed to reproduce our results. Appendix A.1 provides details on the corpus, the held-out splits, the MMseqs2 homology search and thresholds used to remove evaluation homologues. It also states the taxonomic rules and manual corrections defining each exclusion tier, with Table 1 showing the resulting corpus sizes. Appendix A.2 and Table 3 describe the architecture, full pretraining and fine-tuning hyperparameters, and Table 2 lists every size-corpus pair we trained. Appendix A.3 describes the fine-tuning datasets. Appendix A.4 states the recovery criterion, the checkpoint grid, FLOP-counting and cost assumptions behind the dollar figures. Appendix A.5 defines the scoring procedure for each evaluation: the masked-marginals formula and assay counts for mutation effect prediction, as well as the probe, contact definition and PDB filtering criteria (Table 4) for contact prediction.

\section*{Use of Large Language Models}
\label{sec:llm}

Large language models (LLMs) were used for coding, debugging and documentation of the code that supports this work, 
to produce the figures in this paper, to verify all the results against our data and for minor text edits.

\section*{Acknowledgements}
The authors acknowledge the use of resources provided by the Isambard-AI National AI Research Resource (AIRR) \citep{mcintoshsmith2024isambardai}. Isambard-AI is operated by the University of Bristol and is funded by the UK Government's Department for Business, Innovation, Science and Trade (DBIST) via UK Research and Innovation; and the Science and Technology Facilities Council [ST/AIRR/I-A-I/1023].

\bibliography{references}
\bibliographystyle{iclr2026_conference}

\clearpage
\appendix
\raggedbottom
\makeatletter
\setlength{\@fptop}{0pt}
\setlength{\@fpsep}{8pt plus 2pt minus 2pt}
\makeatother
\section*{Appendix}

\section{Extended experimental setup}
\label{sec:app-setup}

\begin{figure}[htbp]
\centering
\includegraphics[width=\textwidth]{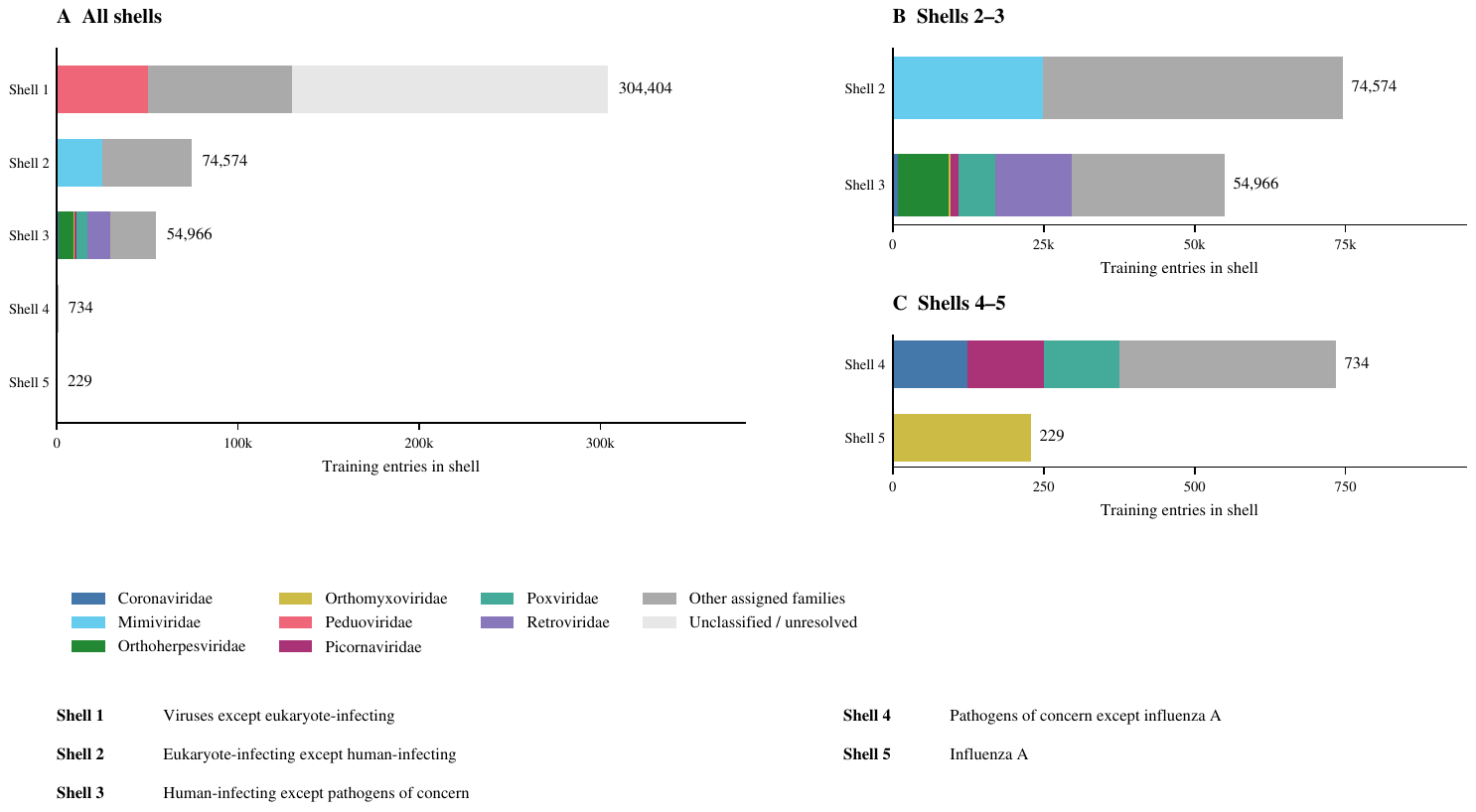}
\caption{\textbf{Distribution of viral families across shells.} Each bar is a shell, and its segments show how that shell's entries are distributed across viral families. This complements the tier-level view in \Cref{fig:exclusion-tiers}.}
\label{fig:appendix-11-shells-by-viral-family}
\end{figure}

\subsection{Pretraining data}

\subsubsection{Corpus and splits}
Following \citet{lin2023esm2}, we use data from UniRef50, 2021\_04 to train our models. We use only sequences from the 49{,}874{,}565 cluster representatives. We exclude 35{,}036 sequences with greater than 10\% \texttt{X} residues, representing unknown amino acids. We then split the remaining sequences into a training set and held-out sets. The held-out set comprises a random sample of 0.5\% of the data ($\approx$ 250{,}000 sequences), which we use for monitoring validation loss during training and an additional set of proteins from each tier shell. For each tier $k \in \{1,\dots,5\}$, we draw a fixed number of sequences from those excluded by tier $k$ but by no narrower tier ($k+1,\dots,5$). This is used in our cross-entropy evaluations. The held-out set is fixed across all models.
MMseqs2 is then used to search the training set for homologues to our tier-specific evaluation set, and to wild-type proteins from ProteinGym and an additional set of 917 \emph{Influenza A} proteins, covering the strains of the mouse-virulence dataset of \citet{ivan2019virulence}, using \texttt{easy-search -c 0.8 -{}-cov-mode 0}. Every training sequence aligning to a query at \texttt{fident} $\geq 0.8$ is dropped. 

The four filtered training sets each remove one tier's sequences from the training set. Membership is decided per row from the cluster representative's taxonomy identifier (\Cref{sec:app-tier-assignment}). Because the exclusion sets are nested, so are the corpora: \texttt{tier0-filtered} $\subset$ \texttt{tier1-filtered} $\subset$ \texttt{tier2-filtered} $\subset$ \texttt{tier3-filtered} $\subset$ \texttt{full}, which we verify row by row on the materialised datasets. \Cref{tab:datasets} gives their sizes.

We run a final search to remove additional \textcolor{forgetorange}{viral} evaluation homologues from the training data, querying the 5{,}161 held-out \textcolor{forgetorange}{viral} sequences, we union \texttt{search -{}-min-seq-id 0.8 -c 0.8 -{}-cov-mode 1} with \texttt{-{}-cov-mode 2}, accepting \texttt{fident} $\geq 0.8$ at $\geq 0.8$ coverage of either sequence. Applied to \textcolor{forgetorange}{viral} sequences only, this leaves a \texttt{full} training set of 49{,}507{,}711. This is run only on \texttt{full}, after the filtered datasets are generated.

\subsubsection{Data labelling and tier assignment}
\label{sec:app-tier-assignment}
NCBI Taxonomy \citep{schoch2020ncbitaxonomy} version 2026-06-01 and Virus-Host DB \citep{mihara2016virushostdb} release 228 are used in determining tier membership.
Tier 1 comprises all sequences whose taxonomy identifier is in the NCBI Taxonomy viral subtree rooted at taxid \texttt{10239}. Since tiers are nested, each additional tier, with the exception of tier 0, is a subset of tier 1. Tier 2 (eukaryote-infecting viruses) comprises every representative whose taxonomy identifier descends from a virus family with at least one eukaryotic host in Virus-Host DB. Tier 3 (human-infecting viruses) applies the same rule with human hosts. Tier 4 (viral pathogens of concern) contains all members of tier 3 (human-infecting) which also appear on at least one of the following biosecurity control lists: the Australia Group's List of Human and Animal Pathogens and Toxins for Export Control (excluding its warning list), or the U.S. HHS (42 CFR 73), USDA Veterinary Services (9 CFR 121) or USDA Plant Protection and Quarantine (7 CFR 331) select-agent lists. Two manual corrections apply: nine taxa on those lists cannot be matched to an NCBI taxon automatically and are added by hand, and one spurious match is removed. We also add Andes, Choclo and Dobrava orthohantavirus, which cause human disease but carry no human host record in Virus-Host DB.

Tier 0 is constructed by aligning every tier 1 viral sequence against the corpus using MMseqs2 with \texttt{--min-coverage 0.8 --cov-mode 0} then selecting sequences with 30\% identity or greater. The tier 0 filter is the union of this and tier 1 sequences.
\Cref{tab:datasets} gives the size of the corpora we pretrain on.

\begin{table}[htbp]
\centering
\caption{\textbf{The pretraining corpora.} Each filtered dataset is the unfiltered UniRef50 corpus with that
tier's entries removed. The tiers are nested, so a wider tier removes everything a narrower one
does. The \texttt{full} dataset underwent an additional filter, removing 9{,}766 additional evaluation homologues after the tier-filtered datasets were constructed.}
\label{tab:datasets}
\begin{tabular}{llrrr}
\toprule
Dataset & Removes & Sequences & Removed & \% Removed \\
\midrule
\texttt{UniRef50 2021\_04} & --- & $49{,}874{,}565$ & --- & --- \\
\midrule
\texttt{full} & --- & $49{,}507{,}711$ & $0$ & 0 \\
\texttt{tier0} filtered & tier 1 $+$ homologues & $47{,}154{,}140$ & $2{,}363{,}337$ & 4.77 \\
\texttt{tier1} filtered & all viruses & $49{,}082{,}570$ & $434{,}907$ & 0.878 \\
\texttt{tier2} filtered & eukaryote-infecting families & $49{,}386{,}974$ & $130{,}503$ & 0.264 \\
\texttt{tier3} filtered & human-infecting families & $49{,}461{,}548$ & $55{,}929$ & 0.113 \\
\bottomrule
\end{tabular}

\end{table}

\subsection{Pre-training}
\label{sec:app-pretraining}
We train models across five sizes: 8M, 35M, 150M, 650M and 3B parameters. The model architecture matches that of ESM-2 as described in \citet{lin2023esm2}. Pretraining hyperparameters are identical across all model sizes, except that gradients are clipped to a maximum norm of 1.0 at 3B; no clipping was applied at other sizes. All models are trained for 150{,}000 steps with a batch size of $\approx$2M tokens. We use the ESM-2 tokeniser from the \texttt{facebook/esm2\_*} Hugging Face checkpoints. Sequences are truncated to 1024 tokens, including \texttt{<cls>} and \texttt{<eos>} tokens, leaving at most 1022 residues. For the masked language-modelling objective, 15\% of residue positions are selected; of these, 80\% are replaced with the mask token, 10\% with a random token, and 10\% are left unchanged, following \citet{devlin2019bert}.

\begin{figure}[htbp]
\centering
\includegraphics[width=\textwidth]{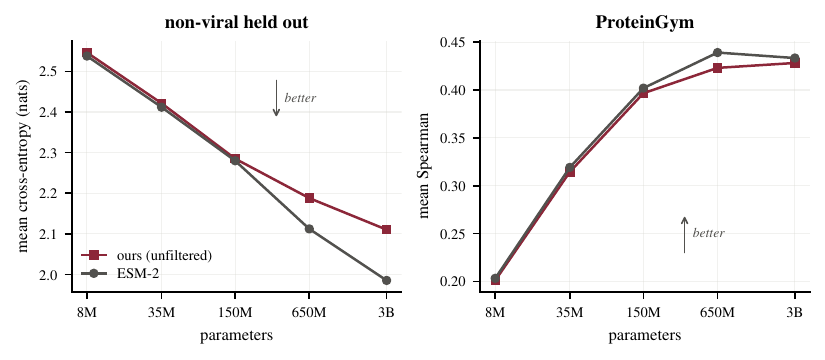}
\caption{\textbf{Our models have similar mutation effect prediction performance to ESM-2 and higher cross-entropy at 650M and 3B parameters.} We compare each unfiltered model with the corresponding \texttt{facebook/esm2} checkpoint. Left: mean cross-entropy over 3{,}000 \textcolor{retainblue}{non-viral} held-out sequences. Right: mean Spearman correlation over 217 ProteinGym substitution assays. These held-out sequences are excluded from our training corpus. ESM-2's training splits are unpublished, so overlap with these sequences is unknown.}
\label{fig:esm2-parity}
\end{figure}

Model weights are randomly initialised and trained using the AdamW optimiser \citep{loshchilov2017adamw} with $\beta_1 = 0.9$, $\beta_2 = 0.98$, a weight decay of $0.01$, and a peak learning rate of $4\times10^{-4}$. The learning rate follows a linear warmup over 2{,}000 steps and a linear decay to $2.93\times10^{-4}$ over the remaining steps. This matches the shape of the ESM-2 schedule, but we stop at 150{,}000 steps rather than the 500{,}000 ESM-2 uses. We chose to train for 150{,}000 steps instead of the 500{,}000 used in \citet{lin2023esm2} due to compute budget. \Cref{fig:esm2-parity} compares our unfiltered model against the stock ESM-2 checkpoint at each size. We find that mutation effect prediction is largely the same across the two models: mean Spearman over the 217 ProteinGym assays is below ESM-2 at every size, but by at most $0.016$. Held-out cross-entropy diverges after 150M, reaching $0.076$ nats at 650M and $0.125$ at 3B. \Cref{tab:models-trained} lists the models we pretrained. We train models at bf16 precision, using the PyTorch \texttt{autocast} feature. For every size, we train on the full corpus and on tier exclusions 0 to 3, with the exception that 3B is only trained on the unfiltered, tier-0 and tier-1 corpora.

\begin{table}[htbp]
\centering
\caption{\textbf{Pretrained models.} A check marks a model size-corpus pair for which a model was pretrained and evaluated. The corpora are the UniRef50 pretraining corpus and the nested tier exclusions of \Cref{fig:exclusion-tiers}.}
\label{tab:models-trained}
\begin{tabular}{lccccc}
\toprule
& \multicolumn{5}{c}{Training corpus} \\
\cmidrule(lr){2-6}
Size & Full & Tier 0 & Tier 1 & Tier 2 & Tier 3 \\
\midrule
8M   & \checkmark & \checkmark & \checkmark & \checkmark & \checkmark \\
35M  & \checkmark & \checkmark & \checkmark & \checkmark & \checkmark \\
150M & \checkmark & \checkmark & \checkmark & \checkmark & \checkmark \\
650M & \checkmark & \checkmark & \checkmark & \checkmark & \checkmark \\
3B   & \checkmark & \checkmark & \checkmark & --- & --- \\
\bottomrule
\end{tabular}
\end{table}

\subsection{Fine-tuning}
\label{sec:app-fine-tuning}
The filtered models are fine-tuned on two different datasets: the 434{,}907 tier-1 viral sequences excluded during tier-1 filtering and, separately, on the same data with Retroviridae sequences removed (leaving 422{,}204 sequences). Each fine-tuning run is initialised from the final pre-training checkpoint, producing two new models from each pretrained checkpoint.
For every fine-tuning run, we use an effective batch of 262{,}144 tokens for 5{,}000 steps, totalling $\approx$1.31B tokens. The learning rate is $5\times10^{-5}$, warmed up linearly over 100 steps and decayed linearly to one tenth of that over the remaining steps. All other hyperparameters are identical to the pretraining setup. \Cref{tab:hyperparameters} gives both settings in full.

\begin{table}[htbp]
\centering
\caption{\textbf{Pretraining and fine-tuning hyperparameters.} Values are shared across model sizes
except where noted. Effective batchsizes are powers of two: $2^{21} = 2{,}097{,}152$ and
$2^{18} = 262{,}144$ tokens. The pretraining schedule decays linearly towards a floor of one tenth of
the peak learning rate at 500{,}000 steps, but the run stops at 150{,}000 steps, where the learning
rate is $2.93\times10^{-4}$. The fine-tuning schedule reaches its floor at the final step.}
\label{tab:hyperparameters}
\begin{tabular}{lll}
\toprule
 & Pretraining & Fine-tuning \\
\midrule
Steps                    & 150{,}000              & 5{,}000 \\
Effective batch          & $2^{21}$ tokens        & $2^{18}$ tokens \\
Total tokens             & $\approx$315B          & $\approx$1.31B \\
Peak learning rate       & $4\times10^{-4}$       & $5\times10^{-5}$ \\
Warmup steps             & 2{,}000                & 100 \\
Decay horizon            & 500{,}000 steps        & 5{,}000 steps \\
Learning-rate floor      & peak${}/10$            & peak${}/10$ \\
Optimiser                & AdamW                  & AdamW \\
$(\beta_1, \beta_2)$     & $(0.9, 0.98)$          & $(0.9, 0.98)$ \\
Weight decay             & $0.01$                 & $0.01$ \\
Gradient clipping        & $1.0$ at 3B only       & $1.0$ at 3B only \\
Maximum sequence length  & 1024 tokens            & 1024 tokens \\
Precision                & bf16 mixed             & bf16 mixed \\
Parallelism              & DDP; FSDP at 3B        & DDP; FSDP at 3B \\
\bottomrule
\end{tabular}
\end{table}

\subsection{Measuring recovery}
\label{sec:app-recovery}

We define performance recovered at the first scored checkpoint where the fine-tuned filtered model reaches
or exceeds the performance of the unfiltered model on the metric in question. Because we score models every 500 steps,
we report the performance as a bracket between the checkpoints. Where a model has already recovered
by the first checkpoint, only the upper edge of the bracket is informative.

We assess per-assay recovery on all 45 EVEREST assays at step 5{,}000 (\Cref{sec:app-assay-recovery}). An assay passes at tolerance $\epsilon$ when its signed Spearman correlation is at least that of the same-size unfiltered model minus $\epsilon$. We also assess recovery per viral family, where a family passes only when every one of its assays passes (\Cref{fig:assay-recovery-tolerance,fig:assay-recovery-scatter,fig:assay-recovery-differences}). For the recovery trajectories (\Cref{fig:assay-recovery-trajectories}), we compare two criteria: every assay matches or exceeds its unfiltered reference, or the equally weighted mean correlation across the family's assays matches or exceeds the corresponding unfiltered mean. We weight the 11 viral families in the EVEREST dataset equally and evaluate the starting models and every 500 steps through step 5{,}000, with 262{,}144 tokens per step. Reaching 80\% passing requires at least 9 families to meet the criterion at the same checkpoint. We recalculate coverage at each checkpoint. 

We count fine-tuning compute as $(6N + 12\,L\,s\,d)\,D$ for $N$ parameters, $L$ layers, width $d$, context
$s=1024$ and $D$ tokens: the usual $6ND$, plus the attention-matrix term that $6ND$ omits. Because the mean
sequence is far shorter than the context, this overcounts the attention term, so the reported cost is an
upper bound.

We convert compute to H100-hours using a conservative estimate of 34\% model-FLOPs utilisation against a 989~TFLOP/s bf16 dense peak (compared to measurements of 38-43\% reported by \citet{grattafiori2024llama3}), and convert H100-hours to dollars at a nominal 4~USD per H100-hour (approximated based on market rates for an eight-H100-SXM instance, at the time of writing). 

Masked-language-modelling recovery at 3B is read from four checkpoints, at 500, 1{,}500, 3{,}000 and
5{,}000 steps, rather than the ten available at other sizes. Its bracket is correspondingly the widest we
report.
\subsection{Evaluation}
\label{sec:app-evaluation}
\subsubsection{Language modelling}
\label{sec:app-language-modelling}
For each sequence $x$, we mask one residue at a time and score the original
residue given the remaining context \citep{salazar2020pseudo}. Let $A(x)$ be the
positions occupied by standard amino acids after truncation to the first
1{,}022 residues; two special tokens give a maximum input length of 1{,}024.
With $x^{(i)}$ denoting the sequence masked at position $i$, we compute
\begin{equation}
 c_\theta(x)=-\frac{1}{|A(x)|}\sum_{i\in A(x)}
 \log p_\theta(x_i\mid x^{(i)}),\qquad
 \mathrm{CE}_\theta(D)=\frac{1}{|D|}\sum_{x\in D}c_\theta(x).
\end{equation}
Every eligible residue is scored; nonstandard residues remain in the context
but are excluded as targets. Sequences receive equal weight irrespective of
length, and individual-sequence pseudo-perplexity is $\exp(c_\theta(x))$.
Standard errors are the sample standard deviation of sequence cross-entropies
divided by $\sqrt{|D|}$; they describe variation across evaluation sequences.
Pooled viral results weight shell means by sequence counts and combine
within-shell and between-shell sums of squares for the standard error.
\subsubsection{Mutation effect prediction}
\label{sec:app-mutation-effect}
We present results from two deep mutational scanning (DMS) benchmarks: ProteinGym \citep{notin2023proteingym} and EVEREST \citep{gurev2025everest}. We score 217 ProteinGym substitution assays, 31 of which are from viral proteins and 186 from non-viral proteins. EVEREST consists of 45 viral assays, 23 of which appear in ProteinGym. We report results from ProteinGym and EVEREST separately rather than in aggregate.

Mutations are scored zero-shot using masked marginals \citep{meier2021esm1v}. For a mutation from wild-type residue $a$ to mutant residue $b$ at position $i$, the score is the log-odds of the mutant relative to the wild type when position $i$ is masked:

\begin{equation}
s(i, a \to b)
= \log p_\theta\!\left(x_i = b \mid x^{(i)}\right)
- \log p_\theta\!\left(x_i = a \mid x^{(i)}\right),
\label{eq:masked-marginal}
\end{equation}

where $x^{(i)}$ denotes the sequence with position $i$ masked. Proteins longer than 1022 residues are cropped to a 1022-residue window centred on the mutation site, following ProteinGym; with the \texttt{<cls>} and \texttt{<eos>} tokens, this matches the 1024-token context used in training.

For multi-site substitutions, we sum the site scores. We report the unweighted
mean of per-assay signed Spearman correlations, retaining negative correlations,
with standard errors across assays. To compare the effect of data filtering on
fitness vs stability prediction, we split ProteinGym and EVEREST by assay type,
reporting ProteinGym viral, ProteinGym non-viral and EVEREST viral results
separately.

\subsubsection{Contact prediction}
\label{sec:app-contact-prediction}
Following \citet{rao2021contacts}, contact prediction is trained over the
attention maps of the frozen model. For residue pair $(i, j)$, the probability of a contact is
\begin{equation}
p(c_{ij}) = \left(1 + \exp\left(-\beta_0 - \sum_{l=1}^{L} \sum_{k=1}^{K} \beta_{kl}\, a^{kl}_{ij}\right)\right)^{-1},
\label{eq:contact-probe}
\end{equation}
where $c_{ij}$ is true when amino acids $i$ and $j$ are in contact. 

$a^{kl}_{ij}$ is the symmetrised and average-product-corrected attention between the pair at head $k$ of layer $l$, and $L$ and $K$ are the model's layer and head counts. We use an L1-regularised ($C = 0.15$) logistic regression probe and train on 20 non-viral sequence chains that contain 100--400 residues, have complete backbone coverage, have a resolution of \(2.0\,\text{\AA}\) or better, and have an assembly size of at most 3.

We use a dataset built from the Protein Data Bank \citep{berman2000pdb}. Starting
from all 552{,}778 protein polymer entities in the mmCIF archive, we keep those of 32 to
1{,}022 residues, containing standard residues only, solved by X-ray crystallography or
cryo-EM at a resolution of 3.0~\AA{} or better, and with at least 80\% of the backbone
resolved, then deduplicate with RCSB's 30\% identity clusters. This leaves 1{,}403
\textcolor{forgetorange}{viral} and 24{,}922 \textcolor{retainblue}{non-viral} cluster
representatives. We sample \textcolor{retainblue}{non-viral} chains to match the number, sequence lengths
and biological-assembly sizes of the \textcolor{forgetorange}{viral} chains, giving 1{,}403 of each. A further 20 chains with fully resolved backbones are held out to fit the probe, giving 2{,}826 chains in total.

As in \citet{rao2021contacts}, we define contacts as pseudo-C$\beta$ distances below 8~\AA{}.
We report $P@L$, $P@L/2$ and $P@L/5$ for short-range ($6 \le j-i < 12$), medium-range ($12 \le j-i < 24$) and long-range ($j-i \ge 24$) pairs, where $j-i$ is the sequence separation. The main text reports long-range $P@L$ only.

\subsubsection{Comparison of degradation between removed and retained viral families}
\label{sec:app-degradation-methods}

\begin{figure}[H]
\centering
\includegraphics[width=\textwidth]{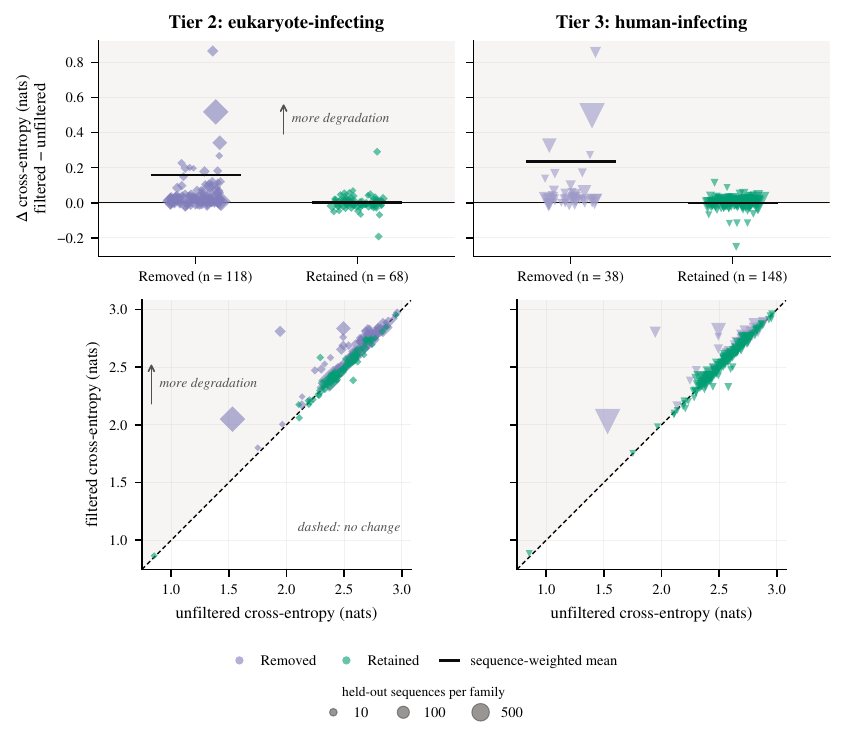}
\caption{\textbf{At 650M, tier-2 and tier-3 filtering increase mean cross-entropy more for removed than retained viral families.} Change in cross-entropy, from the unfiltered model, for 650M models after tier 2 and 3 filtering, split by viral families removed by this filter vs those retained. Top: Mean difference in cross-entropy (filtered minus unfiltered), shown separately for removed and retained families. Each point represents the difference in mean cross-entropy for the held-out sequences for this viral family. Counts in brackets give the number of families. Bars represent sequence-weighted group means. Bottom: Unfiltered versus filtered cross-entropy for retained and removed viral families. The diagonal represents equality.}
\label{fig:degradation-family-comparison}
\end{figure}

\begin{figure}[H]
\centering
\includegraphics[width=\textwidth]{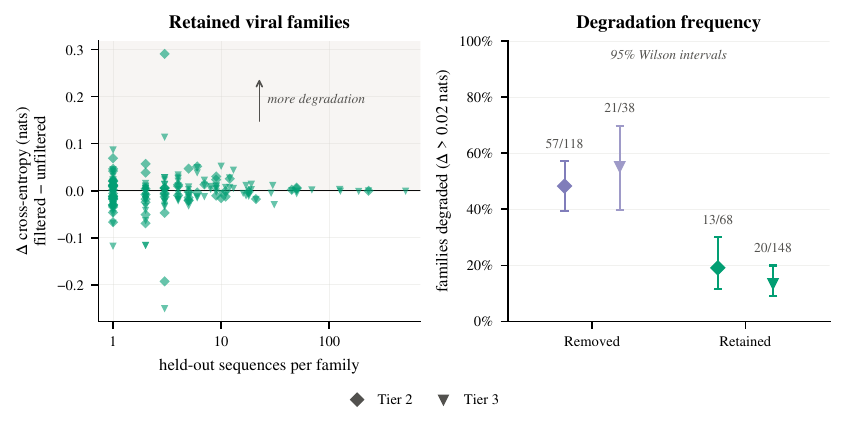}
\caption{\textbf{The largest estimated cross-entropy increases among retained viral families occur in families with few held-out sequences.} Cross-entropy changes for 650M models under tier-2 and tier-3 filtering. Left: Scatter plot showing the change in mean cross-entropy for retained viral families against the number of held-out sequences, highlighting that the largest cross-entropy changes in retained families occur in families with few held-out sequences. Right: Points show the fraction of families that show degradation ($\Delta>0.02$ nats) separately for each tier. Degradation is more frequent among removed than retained families. Error bars show 95\% Wilson intervals, and labels show degraded/total family counts.}
\label{fig:degradation-diagnostics}
\end{figure}

We compare masked-token cross-entropy on held-out \textcolor{forgetorange}{viral} sequences for 650M models
trained on the full corpus and on corpora with tier-2 or tier-3 sequences excluded
(\Cref{fig:degradation-family-comparison,fig:degradation-diagnostics}). We define
$\Delta=\mathrm{CE}_{\mathrm{excluded}}-\mathrm{CE}_{\mathrm{full}}$ in nats, so positive
values indicate worse predictions after filtering. Tier 2 excludes eukaryote-infecting
\textcolor{forgetorange}{viral} families; Tier 3 excludes human-infecting \textcolor{forgetorange}{viral} families.

We assign training and held-out sequences to families using NCBI Taxonomy
version 2026-09-01, excluding unclassified \textcolor{forgetorange}{viral} sequences. A family is removed
if it has no sequences in the tier's training corpus and retained otherwise.
We average each model's cross-entropy and $\Delta$ within families over the same
held-out sequences. In \Cref{fig:degradation-diagnostics}, points show the proportion
of families with mean $\Delta>0.02$ nats, with 95\% Wilson intervals ($z=1.96$)
reported separately for each tier and group.

Family means estimated from few held-out sequences are noisy. Models trained on
corpora with tier-2 or tier-3 sequences excluded were limited to 650M parameters.
As \textcolor{forgetorange}{viral} families recur across tiers, the per-tier estimates are not independent.

\subsection{Measuring recovery excluding Retroviridae}
\label{sec:app-retroviral-methods}

In \Cref{fig:retroviral-appendix,fig:retroviral-trajectories,fig:retroviral-recovery-cost},
we fine-tune tier-0 and tier-1 models from \Cref{sec:app-pretraining} on the viral
corpus with Retroviridae removed, using the procedure and hyperparameters in
\Cref{sec:app-fine-tuning,tab:hyperparameters}. We compare each trajectory with its same-size
unfiltered pretrained reference, using the evaluation methods in
\Cref{sec:app-evaluation} and recovery criterion and compute conversions in
\Cref{sec:app-recovery}. 

For the human-infecting sequence evaluation in
\Cref{fig:retroviral-trajectories,fig:retroviral-recovery-cost}, removing 926
Retroviridae entries leaves 535 sequences. We score each fine-tuned checkpoint,
its pretrained origin and its unfiltered reference on these same identifiers.
Other sequence strata are unchanged. EVEREST retains all 45 assays, including
the three Retroviridae assays, and uses the signed-Spearman mean defined in
\Cref{sec:app-mutation-effect}. The scatter in \Cref{fig:retroviral-appendix}
pairs individual-sequence pseudo-perplexities for the 3B unfiltered and tier-0
pretrained models, with disjoint nonviral, other-viral and Retroviridae groups.

\clearpage

\section{Extended results}
\label{sec:app-results}

\subsection{Removed and retained viral families}
\label{sec:app-degradation}

\subsection{Contact Prediction}

We measure the filtering effect and recovery across model sizes. We evaluate precision separately for each chain of the 1{,}403 viral and 1{,}403 matched non-viral PDB chains described in \Cref{sec:app-contact-prediction}. While the main body reports P@L as our metric for long-range precision (sequence separation $\geq 24$ residues), here we also report P@L/2 and P@L/5, for short- and medium-range pairs as well as long-range (ranges defined in \Cref{sec:app-contact-prediction}).

\begin{figure}[H]
\centering
\includegraphics[width=0.95\textwidth]{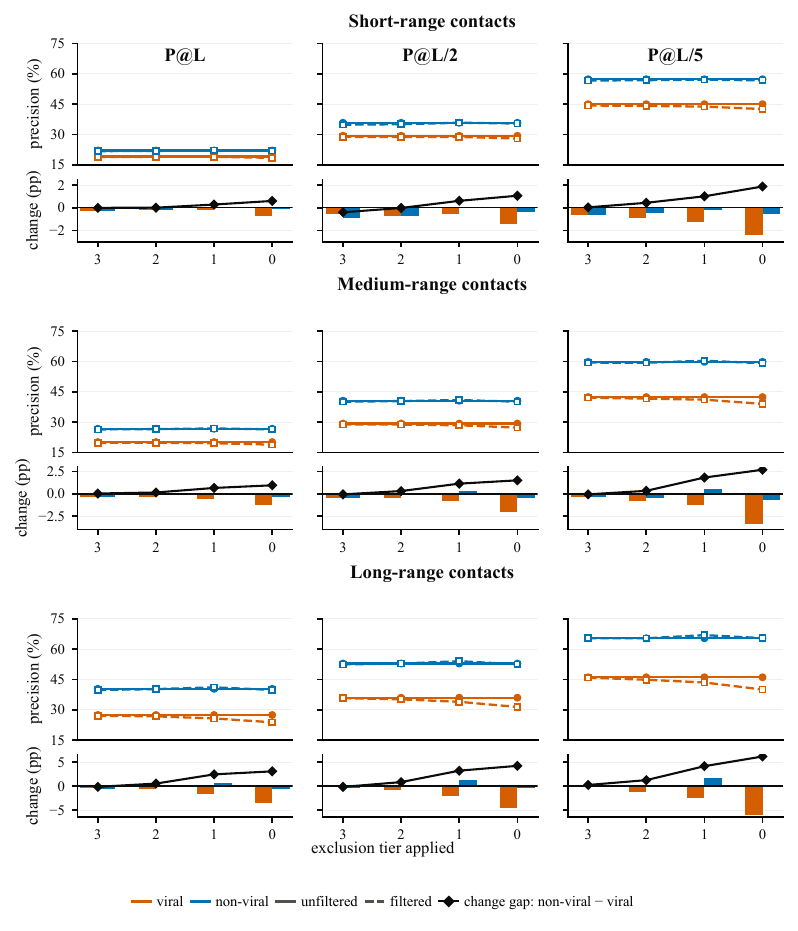}
\caption{\textbf{Filtering effect on contact prediction in short-, medium-, and long-range residue-residue pairs for the 650M model}. Each row compares P@L, P@L/2, and P@L/5 across four filtering levels, from Tier 3 to Tier 0, against the unfiltered model. The upper panels show mean precision for viral and non-viral chains before and after filtering; the lower panels show each domain's change (filtered minus unfiltered) and the non-viral minus viral change gap, in percentage points. }
\label{fig:contact-filtering-levels}
\end{figure}

We evaluate 650M models across filtering levels and measure precision (P@L, P@L/2, and P@L/5) for all three contact ranges (short, medium, and long; \Cref{fig:contact-filtering-levels}). Stronger filtering results in a larger loss of contact-prediction precision on viral chains. For long-range P@L, the viral change relative to the unfiltered model falls from $-0.41$ percentage points at Tier 3 to $-3.60$ at Tier 0; the mean within-chain slope is $-1.06$ percentage points per ordered filtering level (95\% CI $[-1.23,-0.90]$; two-sided one-sample $t$-test, $p=5.4\times10^{-35}$). Non-viral changes across the four tiers range from $-0.57$ to $+0.73$ percentage points. The non-viral-minus-viral change gap also increases with filtering strength: for long-range P@L, its mean slope is $+1.16$ percentage points per ordered level (95\% CI $[0.98,1.34]$; two-sided Welch test, $p=2.3\times10^{-35}$). This overall gap trend is significant in all nine range--precision combinations (BH-adjusted $q < 0.05$ under both Welch and Mann--Whitney tests). The relative effect is largest for long-range contacts and the highest-ranked predictions (P@L/5).

\begin{figure}[H]
\centering
\includegraphics[width=0.95\textwidth]{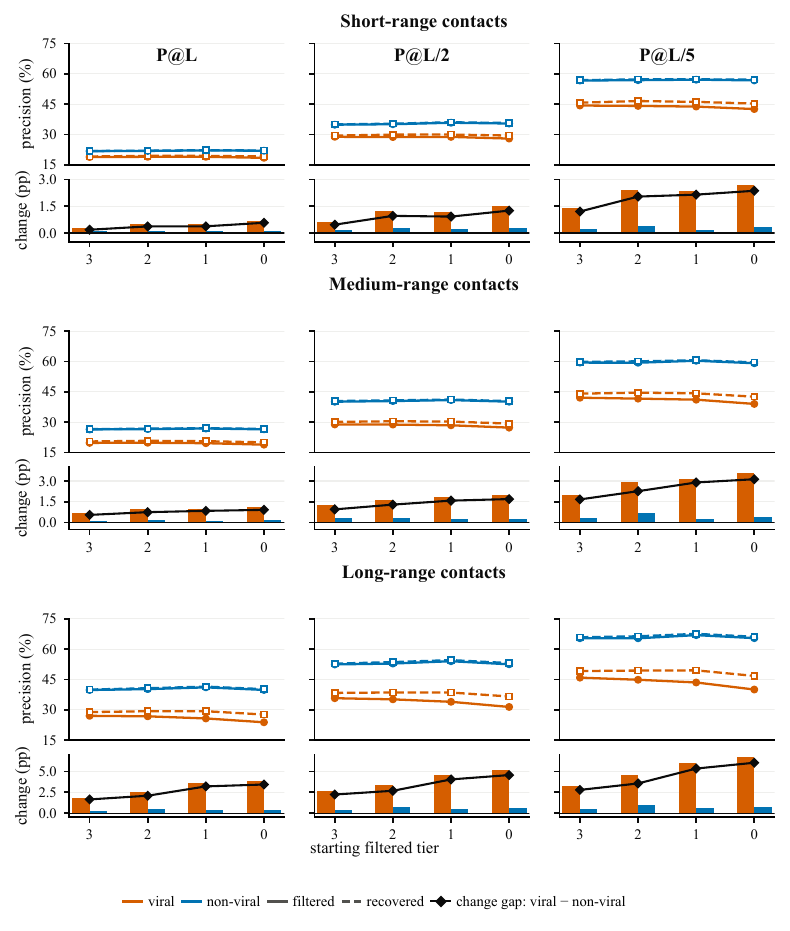}
\caption{\textbf{fine-tuning recovers contact prediction performance across short-, medium-, and long-range residue-residue pairs in the 650M model}. Each row compares P@L, P@L/2, and P@L/5 for filtered checkpoints from Tier 3 to Tier 0 with the corresponding models after viral-only fine-tuning (see \Cref{sec:app-fine-tuning}). The upper panels show mean precision for viral and non-viral chains before and after fine-tuning; the lower panels show each domain's change (recovered minus filtered) and the viral minus non-viral change gap, in percentage points.}
\label{fig:recovery-levels}
\end{figure}

To measure the recovery effect, we compare 650M models after 5{,}000 steps of viral-only fine-tuning with their corresponding filtered starting checkpoints (tiers 3--0; \Cref{sec:app-fine-tuning}), using P@L, P@L/2, and P@L/5 across all three contact ranges on viral and non-viral chains (\Cref{fig:recovery-levels}). For long-range P@L, viral gains rise from $+1.85$~pp at tier 3 to $+3.83$~pp at tier 0, while non-viral gains across the four tiers range from $+0.21$ to $+0.44$~pp. The viral-minus-non-viral gaps are $+1.65$, $+2.09$, $+3.21$, and $+3.44$~pp at tiers 3, 2, 1, and 0, respectively. Viral recovery exceeds non-viral recovery in all 36 tier--range--precision comparisons (BH-adjusted $q<0.05$ under both one-sided Welch and Mann--Whitney tests). The gap increases overall as the starting filter strengthens from tier 3 to tier 0 ($+0.65$~pp per ordered level, 95\% CI $[0.48,0.82]$; one-sided Welch $p=1.3\times10^{-14}$, BH-adjusted $q=1.1\times10^{-13}$). This overall trend is supported in all nine contact-range $\times$ precision combinations (BH-adjusted $q<0.05$ under both one-sided Welch and Mann--Whitney tests). The preferential recovery effect is largest for long-range contacts and the highest-ranked predictions (P@L/5).

\begin{figure}[H]
\centering
\includegraphics[width=0.95\textwidth]{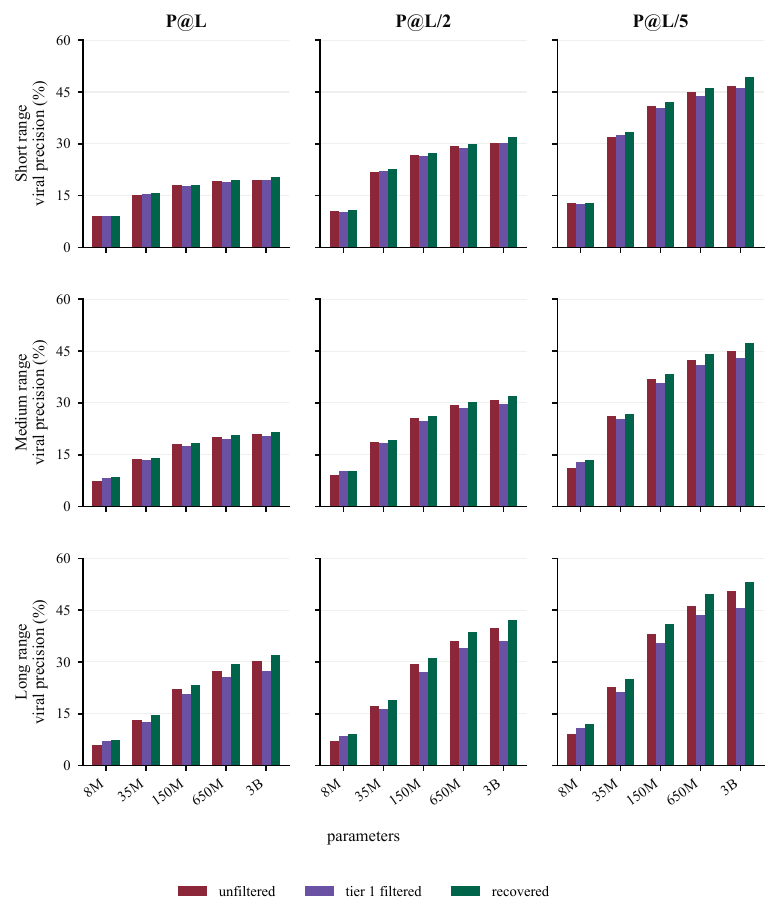}
\caption{\textbf{Viral contact-prediction precision across model sizes before filtering, after tier-1 filtering, and after viral-only fine-tuning}. Rows show short-, medium-, and long-range contacts; columns show P@L, P@L/2, and P@L/5. Each panel compares the mean precision of the unfiltered, tier-1 filtered, and recovered models on viral chains.}
\label{fig:recovery-overview}
\end{figure}

After viral-only fine-tuning from tier 1, mean viral precision increases relative to the filtered starting checkpoint in all 45 size--range--precision combinations, recovering to or surpassing the same-size unfiltered level in 44 of 45 (\Cref{fig:recovery-overview}). All 15 P@L recovered-minus-unfiltered gains are significant in paired comparisons of the 1{,}403 viral chains (BH-adjusted $q<0.05$ under both two-sided paired $t$ and Wilcoxon signed-rank tests, correcting across all 45 combinations). For long-range P@L at 650M, the gain over unfiltered is $+1.87$~pp (95\% CI $[1.52,2.22]$; two-sided paired $t$-test $p=1.7\times10^{-24}$, BH-adjusted $q=8.6\times10^{-24}$). Across all metrics, 43 of 45 recovered-minus-unfiltered differences are positive and significant under both tests; short-range P@L/2 at 8M is positive but not significant, and short-range P@L/5 at 8M has a $-0.16$~pp mean difference.

\begin{figure}[H]
\centering
\includegraphics[width=0.95\textwidth]{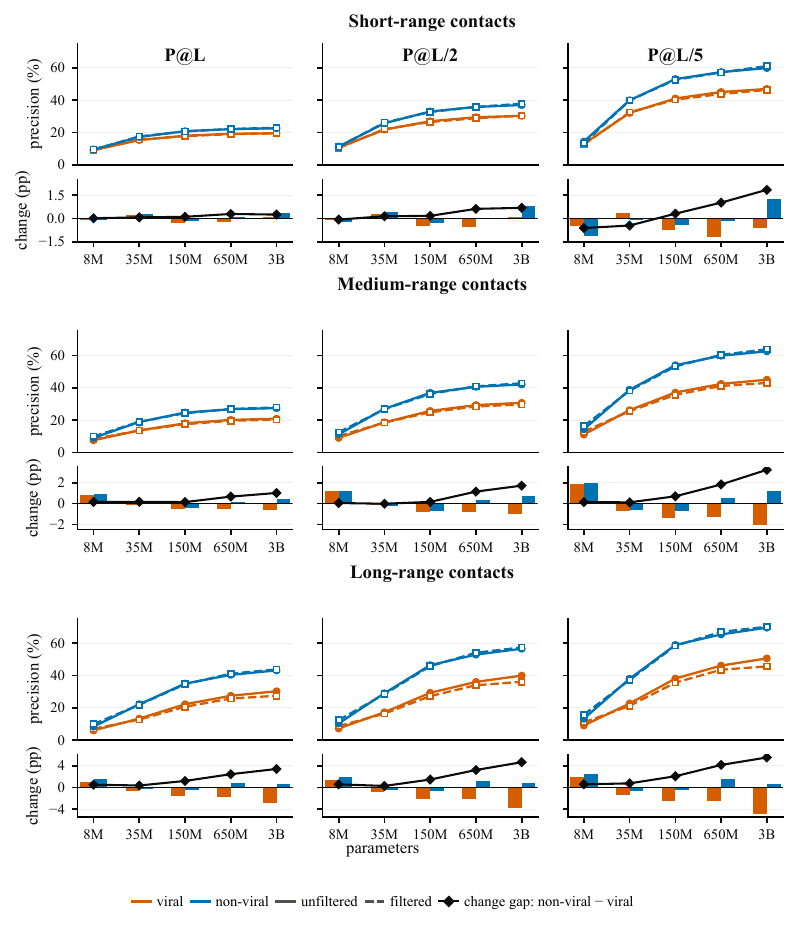}
\caption{\textbf{Scaling of the filtering effect in short-, medium-, and long-range pairs}. Each row compares unfiltered and tier-1 filtered models from 8M to 3B using P@L, P@L/2, and P@L/5. The upper panels show mean precision for viral and non-viral chains; the lower panels show each domain's change (filtered minus unfiltered) and the non-viral minus viral change gap, in percentage points. The gap is larger for medium- than short-range pairs at 650M and 3B, and larger for long- than medium-range pairs at 150M--3B, across all three metrics.}
\label{fig:filtering-scaling}
\end{figure}

To test whether the filtering effect grows with model size, we compare unfiltered and tier-1 filtered models at five sizes (\Cref{fig:filtering-scaling}). For each chain, we calculate the change in precision (filtered minus unfiltered) at each of the five sizes and estimate its slope against $\log_{10}$ parameter count by ordinary least squares. The non-viral-minus-viral slope difference is positive in all nine contact-range $\times$ precision combinations (BH-adjusted $q<0.05$ under both two-sided Welch and Mann--Whitney tests); for long-range P@L, it is $+1.23$~pp per tenfold increase in parameter count (95\% CI $[1.00,1.45]$; Welch $q=2.1\times10^{-25}$, Mann--Whitney $q=2.1\times10^{-17}$). Viral precision loss itself grows with size in seven of nine combinations (BH-adjusted $q<0.05$ under both two-sided one-sample $t$ and Wilcoxon signed-rank tests), with short-range P@L and P@L/2 as the exceptions.

\begin{figure}[H]
\centering
\includegraphics[width=0.95\textwidth]{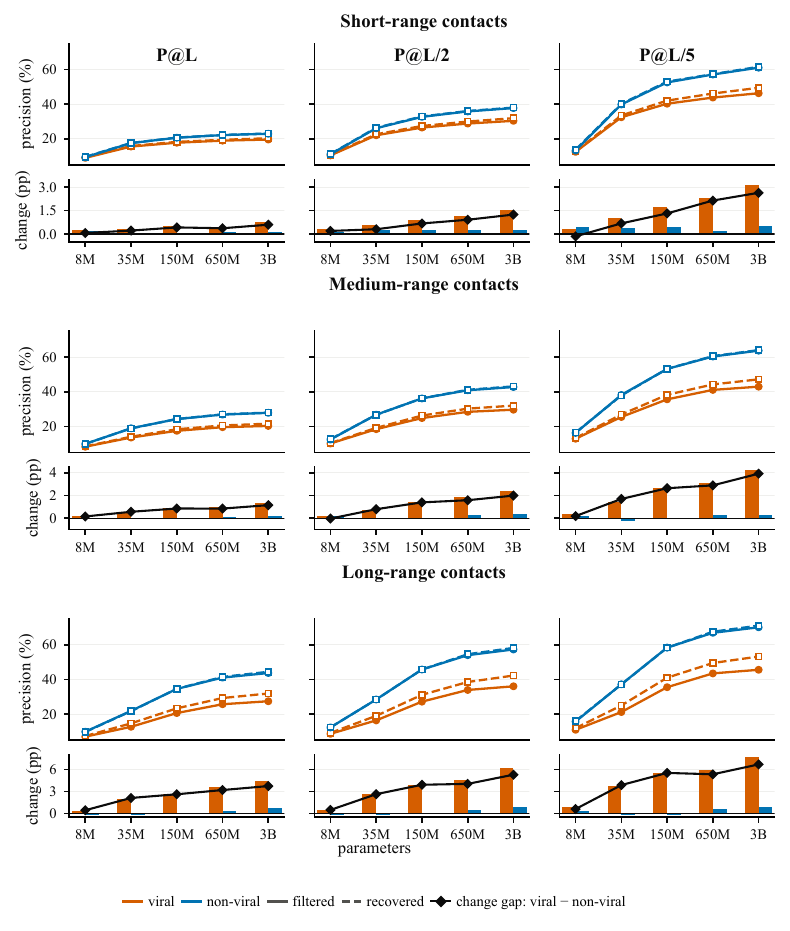}
\caption{\textbf{Scaling of the recovery effect in short-, medium-, and long-range pairs}. Each row compares tier-1 filtered models from 8M to 3B with their counterparts after viral-only fine-tuning using P@L, P@L/2, and P@L/5. The upper panels show mean precision for viral and non-viral chains before and after fine-tuning; the lower panels show each domain's change (recovered minus filtered) and the viral minus non-viral change gap, in percentage points. For P@L, the gap increases from short to medium to long range at every model size.}
\label{fig:recovery-scaling}
\end{figure}

To test whether the recovery effect grows with model size, we compare each tier-1 filtered model with its counterpart after viral-only fine-tuning (\Cref{fig:recovery-scaling}). For each chain, we calculate the change in precision (recovered minus filtered) at each of the same five sizes and estimate its slope against $\log_{10}$ parameter count by ordinary least squares. The viral-minus-non-viral slope difference is positive in all nine contact-range $\times$ precision combinations (BH-adjusted $q<0.05$ under both two-sided Welch and Mann--Whitney tests); for long-range P@L, it is $+1.19$~pp per tenfold increase in parameter count (95\% CI $[0.99,1.39]$; Welch $q=3.4\times10^{-30}$, Mann--Whitney $q=3.0\times10^{-15}$). Viral recovery itself grows with size in all nine combinations (BH-adjusted $q<0.05$ under both two-sided one-sample $t$ and Wilcoxon signed-rank tests). These tests measure variation across chains for the evaluated checkpoints, not across independently trained model runs.

\subsection{Filtering effects by viral family}

\begin{figure}[htbp]
\centering
\includegraphics[width=\textwidth]{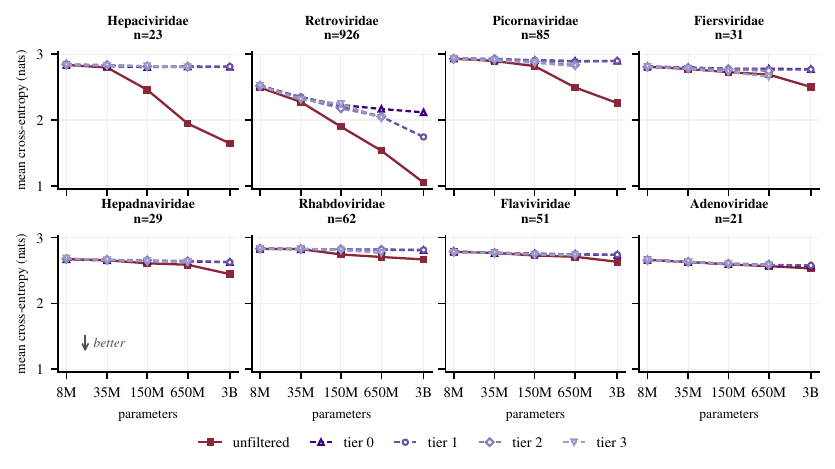}
\caption{\textbf{Filtering differentially degrades language modelling performance across proteins from different viral families.}
Each plot shows mean cross-entropy on held-out entries for the eight families with the largest tier-1 effect at 3B (among families with at least
20 entries). Each curve shows the unfiltered and tier 0 to tier 3 models for each model size up to 650M, and tier 0 and 1 for 3B.}
\label{fig:family-scaling}
\end{figure}

A small number of \textcolor{forgetorange}{viral} families account for much of the change in cross-entropy observed after filtering (\Cref{fig:family-scaling}). For those \textcolor{forgetorange}{viral} families with at least 20 sequences in our test set, we find that in the tier-1 filtered 3B model, Retroviridae accounts for $\approx$~77.5\% of the summed increase in held-out cross-entropy (weighted by sequence count). This is followed by Picornaviridae ($\approx$~6.56\%), unclassified viruses ($\approx$~4.11\%), Hepaciviridae ($\approx$~3.25\%) and Mimiviridae ($\approx$~1.17\%).

\begin{table}[htbp]
\centering
\caption{\textbf{Effects of filtering on masked language-modelling and mutation effect prediction by model size.} Left: Mean paired differences between tier 1 filtered and unfiltered models in cross-entropy (nats, 5{,}161 held-out viral sequences). Positive values indicate that the tier 1 filtered model performs better. Middle: Change in mean Spearman correlation for mutation-effect prediction (Spearman $\rho$, 31 assays) for tier 1 - unfiltered. Negative values indicate that the tier 1 filtered model performs worse. Right: Difference in cross-entropy (nats) between tier 0 and tier 1 filtered models on paired sequences. $p$-values are calculated from two-sided paired $t$-tests, without adjustment for multiple comparisons.}
\label{tab:significance}
\begin{tabular}{lrrrrrr}
\toprule
& \multicolumn{2}{c}{Cross-entropy, Tier 1} & \multicolumn{2}{c}{ProteinGym, Tier 1} & \multicolumn{2}{c}{Tier 0 $-$ Tier 1} \\
\cmidrule(lr){2-3}\cmidrule(lr){4-5}\cmidrule(lr){6-7}
Size & $\Delta$ (nats) & $p$ & $\Delta\rho$ & $p$ & $\Delta$ (nats) & $p$ \\
\midrule
8M & $+0.006$ & $4\times 10^{-35}$ & $+0.001$ & $0.844$ & $+0.0005$ & $0.330$ \\
35M & $+0.019$ & $<10^{-99}$ & $-0.002$ & $0.816$ & $+0.0007$ & $0.269$ \\
150M & $+0.074$ & $4\times 10^{-81}$ & $-0.019$ & $0.264$ & $+0.0121$ & $9\times 10^{-42}$ \\
650M & $+0.125$ & $<10^{-99}$ & $-0.068$ & $0.003$ & $+0.0362$ & $<10^{-99}$ \\
3B & $+0.173$ & $<10^{-99}$ & $-0.124$ & $3\times 10^{-4}$ & $+0.0855$ & $<10^{-99}$ \\
\bottomrule
\end{tabular}

\end{table}

\begin{table}[htbp]
\centering
\caption{\textbf{Per-shell effects of filtering at 650M}
Left: mean paired cross-entropy differences
(tier 1 minus unfiltered, nats) and the share for each shell
of the summed difference across all five viral shells.
Shares are weighted by the number of held-out sequences
in each shell.
Right: mean paired cross-entropy differences
(tier 0 minus tier 3, nats).
Both comparisons pair observations by sequence.
$p$-values are from two-sided paired $t$-tests,
without adjustment for multiple comparisons.}
\label{tab:shells}
\begin{tabular}{lrrrrr}
\toprule
& & \multicolumn{2}{c}{Tier 1 vs unfiltered} & \multicolumn{2}{c}{Tier 0 $-$ Tier 3} \\
\cmidrule(lr){3-4}\cmidrule(lr){5-6}
Shell & $n$ & $\Delta$ (nats) & Share & $\Delta$ (nats) & $p$ \\
\midrule
1 -- phages, unclassified & 1462 & $+0.029$ & 6.5\% & $+0.068$ & $5\times 10^{-61}$ \\
2 -- eukaryote-infecting & 1472 & $+0.027$ & 6.2\% & $+0.038$ & $1\times 10^{-40}$ \\
3 -- human-infecting & 1461 & $+0.361$ & 81.6\% & $+0.098$ & $<10^{-99}$ \\
4 -- pathogens of concern & 493 & $+0.070$ & 5.3\% & $+0.015$ & $2\times 10^{-5}$ \\
5 -- influenza A & 273 & $+0.008$ & 0.3\% & $+0.007$ & $0.003$ \\
\bottomrule
\end{tabular}

\end{table}

\begin{figure}[htbp]
\centering
\includegraphics[width=\textwidth]{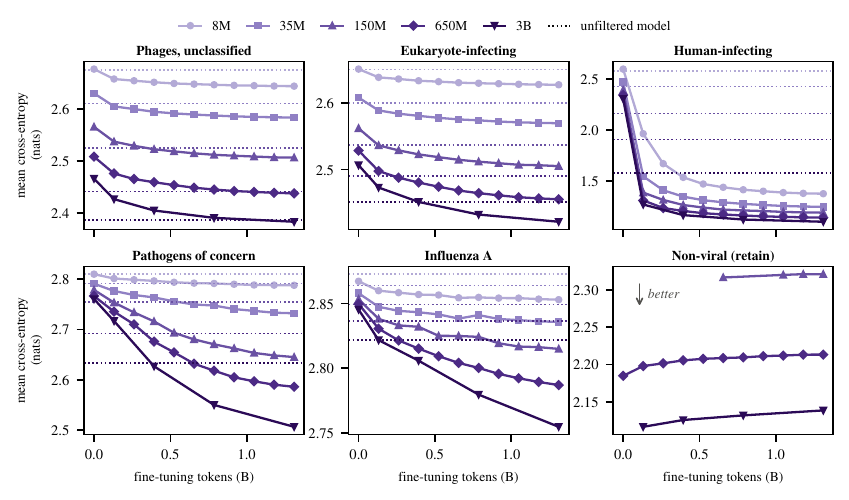}
\caption{\textbf{Viral cross-entropy decreases during fine-tuning, while non-viral cross-entropy increases over the evaluated checkpoints at 650M and 3B.}
Mean cross-entropy (nats) plotted against fine-tuning tokens (billions) for tier-0-filtered models trained on the withheld \textcolor{forgetorange}{viral} corpus.
Panels show language modelling performance on the five viral shells and the \textcolor{retainblue}{non-viral} control.
The dotted horizontal lines show the performance of the unfiltered baseline at each model size.
Lower cross-entropy indicates better performance.
Non-viral cross-entropy increases over the evaluated checkpoints at 650M and 3B.}
\label{fig:recovery-all-shells}
\end{figure}

\begin{figure}[htbp]
\centering
\includegraphics[width=\textwidth]{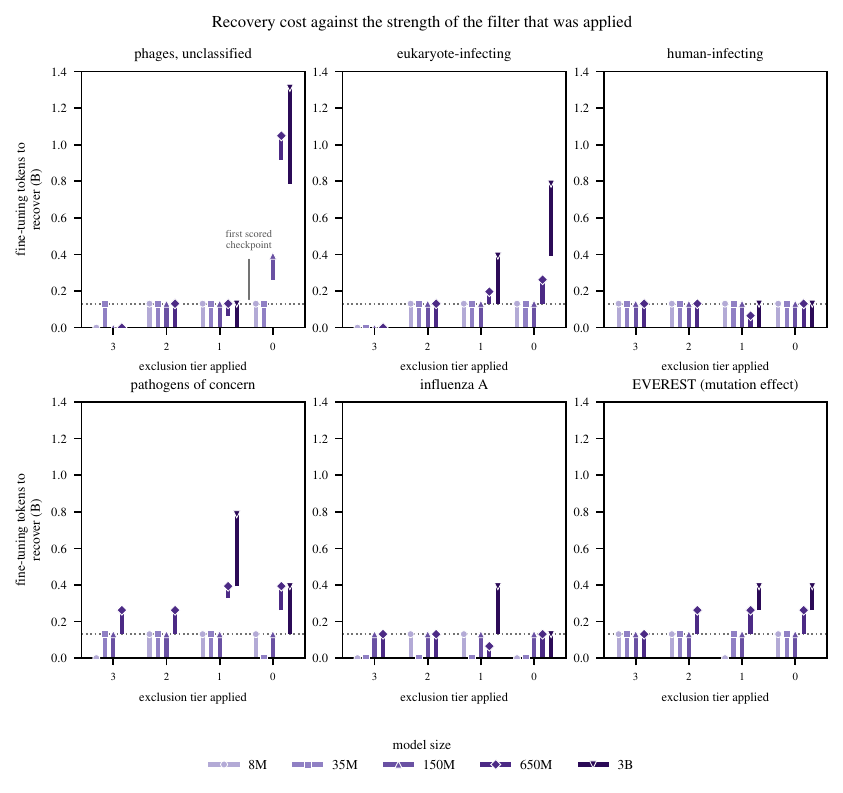}
\caption{\textbf{Mean cross-entropy on the human-infecting shell recovers within 1.4B fine-tuning tokens for every evaluated model size and filtering tier.}
Recovery of mean cross-entropy on the five viral shells and mean Spearman correlation on EVEREST during viral-only fine-tuning.
Exclusion tiers run from tier 3 (narrowest) to tier 0 (widest).
Recovery is defined as matching or improving on the mean performance of the same-size unfiltered model.
Vertical bars show the bracket of recovery between scored checkpoints, where upper endpoints give the first checkpoint meeting this criterion.
Intervals starting at zero indicate recovery before the first scored fine-tuning checkpoint.
The dotted line marks 500 fine-tuning steps (approximately 0.131 billion tokens).
All available size and filter combinations recover on the human-infecting shell within this fine-tuning budget.
No tier-2 or tier-3 filtered model was trained at 3B.}
\label{fig:recovery-by-tier}
\end{figure}

\begin{figure}[htbp]
\centering
\includegraphics[width=\textwidth]{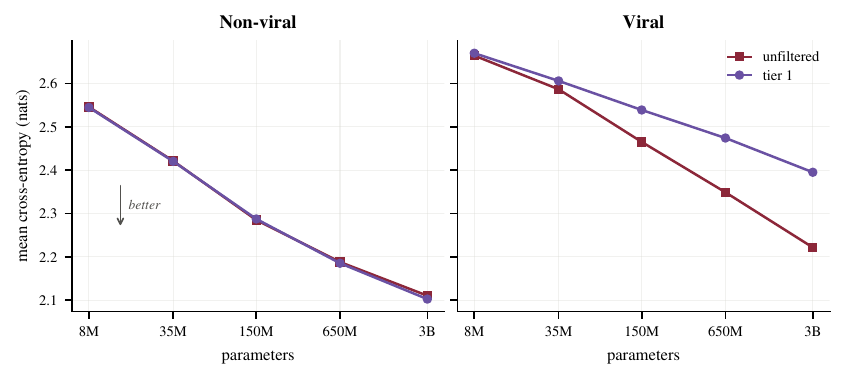}
\caption{\textbf{The increase in viral cross-entropy from tier-1 filtering grows with model size.}
Mean held-out cross-entropy (nats) for unfiltered and tier-1-filtered models from 8M to 3B parameters, evaluated on \textcolor{retainblue}{non-viral} (left) and \textcolor{forgetorange}{viral} (right) sequences. Lower cross-entropy indicates better performance.
\textcolor{forgetorange}{Viral} cross-entropy decreases with model size under both training conditions, with a larger reduction for the unfiltered model.
The filtered-minus-unfiltered performance difference increases from $0.006$~nats at 8M to $0.173$~nats at 3B. Differences in performance on \textcolor{retainblue}{non-viral} remain within $\pm 0.008$~nats at all five sizes.
Comparisons across filtering tiers are shown in \Cref{fig:exclusion-cost}.}
\label{fig:ppl-forget-retain}
\end{figure}

\begin{figure}[htbp]
\centering
\includegraphics[width=\textwidth]{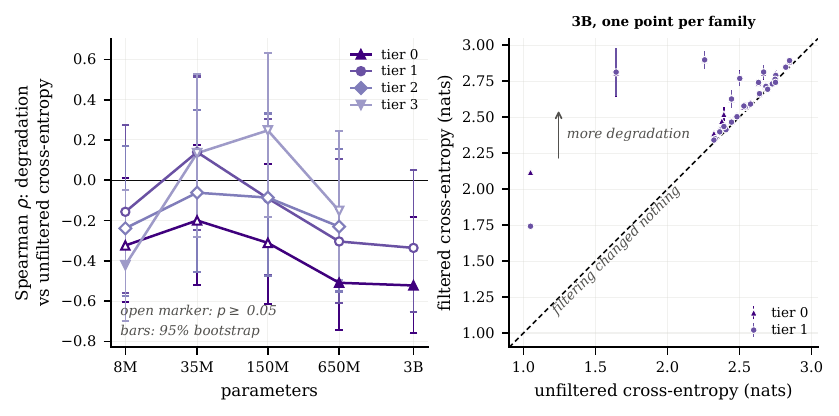}
\caption{\textbf{Tier-0 degradation is negatively correlated with unfiltered cross-entropy at 650M and 3B.}
Association between unfiltered cross-entropy and the change in cross-entropy after filtering across viral families.
Left: Spearman correlation across the 30 families (including unclassified) with at least 20 held-out entries.
When performing the statistical analysis, each family's entries are split into two disjoint halves
with the same split used across model sizes and filtering tiers. One half estimates unfiltered cross-entropy and the other estimates the paired filtered-minus-unfiltered difference, so each estimate is calculated from a separate set of held-out sequences.
Bars are 95\% percentile intervals from 10{,}000 family bootstrap resamples.
Open markers indicate $p \geq 0.05$ in a two-sided test with 10{,}000 permutations of family pairings.
Intervals and tests are conditional on the fixed sequence split, and $p$-values are not adjusted for multiple comparisons.
Lower cross-entropy is better; therefore, a negative correlation suggests that greater degradation due to filtering occurs in families where the unfiltered performance is higher.
Right: filtered versus unfiltered cross-entropy at 3B, using all held-out entries in each family.
Each point is a \textcolor{forgetorange}{viral} family. Dashed line marks equal performance and vertical bars show the paired standard error of the difference.}
\label{fig:family-correlation}
\end{figure}

\begin{figure}[htbp]
\centering
\includegraphics[width=\textwidth]{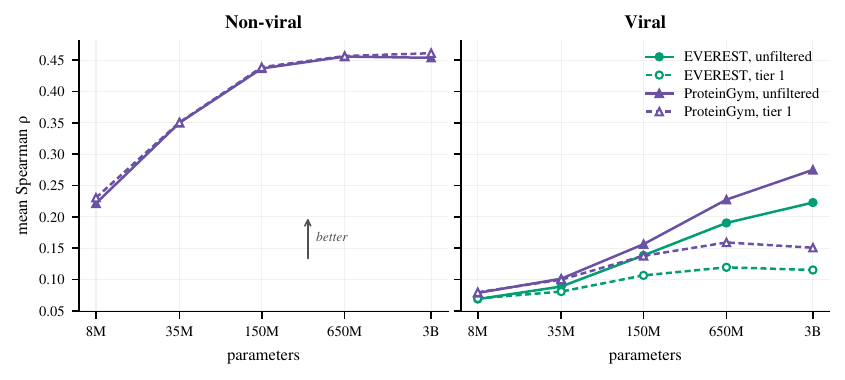}
\caption{\textbf{Mean Spearman correlation on viral assays increases more from 150M to 3B in unfiltered models than in tier 1 filtered models.}
Mean per-assay Spearman $\rho$ between predicted and measured mutation effects, for unfiltered and tier 1 filtered models. Left: 186 \textcolor{retainblue}{non-viral} ProteinGym assays.
Right: 31 \textcolor{forgetorange}{viral} ProteinGym assays and 45 EVEREST assays (all EVEREST assays are \textcolor{forgetorange}{viral}). On \textcolor{forgetorange}{viral} assays, unfiltered models show
larger gains in mean correlation than tier 1 filtered models from 150M onwards.
On \textcolor{retainblue}{non-viral} assays, the difference in mean correlation between filtering conditions is less than 0.009
in magnitude at every model size.}
\label{fig:dms-scaling}
\end{figure}

\begin{figure}[htbp]
\centering
\includegraphics[width=\textwidth]{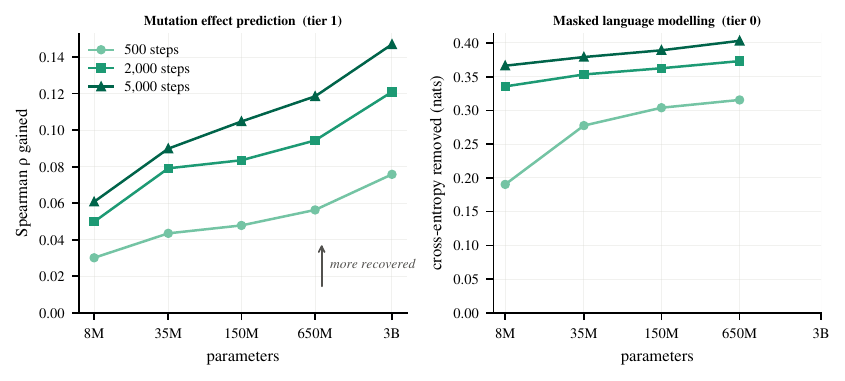}
\caption{\textbf{Larger filtered models show greater performance gains after the same number of fine-tuning steps.} Performance recovered
after a fixed number of fine-tuning steps for different model sizes. Changes are measured from each model's filtered starting checkpoint after 500, 2,000 and 5,000 fine-tuning steps. Left: Increase in mean Spearman $\rho$ on EVEREST for tier 1 filtered models. Right: Decrease in mean held-out viral cross-entropy (nats) for tier 0 filtered models up to 650M. The gain rises with model size at all three budgets. The right panel stops at 650M as this analysis was not performed for 3B.}
\label{fig:sample-efficiency}
\end{figure}

\begin{figure}[htbp]
\centering
\includegraphics[width=\textwidth]{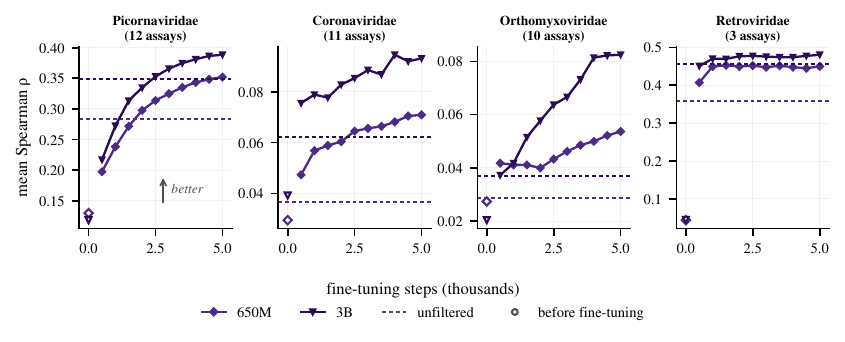}
\caption{\textbf{At 650M and 3B, mean Spearman correlation exceeds the unfiltered reference at 2{,}500 fine-tuning steps in all four families with at least three EVEREST assays.}
Mean per-assay Spearman $\rho$ between predicted and measured mutation effects for 650M and 3B tier 1 filtered models
during fine-tuning on the withheld \textcolor{forgetorange}{viral} dataset. Panels show the four \textcolor{forgetorange}{viral} families with at least
three EVEREST assays. Note that vertical scales differ between panels.
Fine-tuning steps are shown in thousands. Dashed lines indicate the mean correlation for the same-size unfiltered model while
open markers indicate the filtered model before fine-tuning. At both model sizes, the mean correlation for each displayed
family is below the corresponding unfiltered reference before fine-tuning and exceeds it at the 2{,}500-step checkpoint.}
\label{fig:recovery-everest}
\end{figure}

\clearpage
\subsection{Recovery after fine-tuning without Retroviridae}
\label{app:retroviral-recovery}

\begin{figure}[htbp]
\centering
\includegraphics[width=\textwidth]{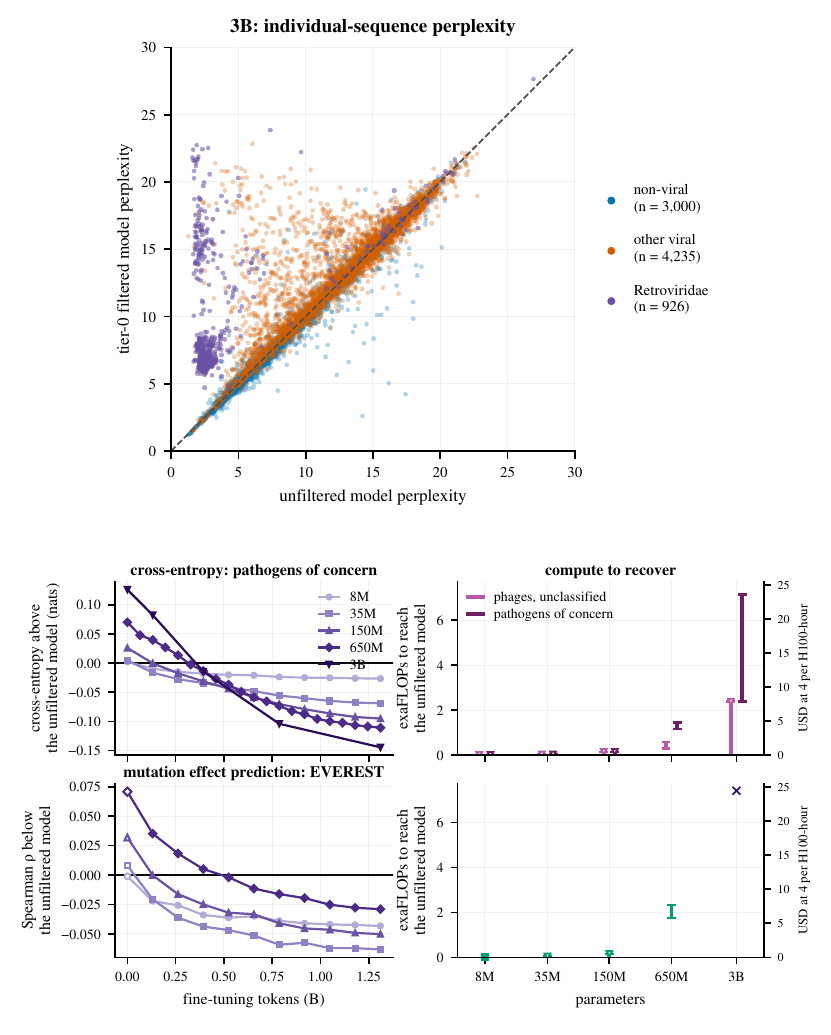}
\caption{\textbf{Sequence perplexity and recovery without Retroviridae.}
Top: matched 3B unfiltered and tier-0 pretrained models, with one point per sequence,
with 3{,}000 nonviral, 926 Retroviridae and 4{,}235 other viral sequences. 
Dashed line marks equal pseudo-perplexity. 
Bottom: tier-1 starting models fine-tuned without Retroviridae,
cross-entropy and mean signed Spearman gaps to the same-size unfiltered model
(positive means worse), with observed recovery-compute brackets.
The dollar axis assumes 34\% H100 utilisation and \$4 per H100-hour.
EVEREST averages all 45 assays. Recovery brackets span adjacent evaluations;
open cost markers denote first-checkpoint upper bounds, squares at zero denote
initial parity, and crosses denote unavailable results. Open trajectory markers
at zero tokens identify pretrained starting models.}
\label{fig:retroviral-appendix}
\end{figure}

\begin{figure}[p]
\centering
\includegraphics[width=\textwidth]{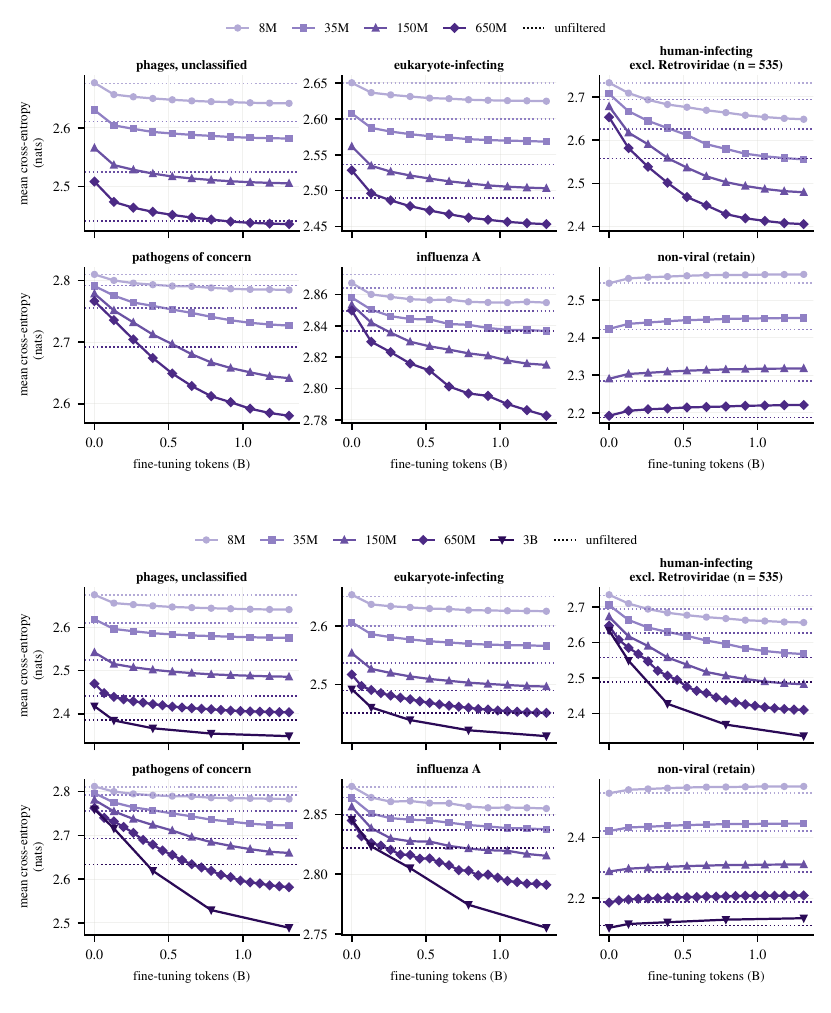}
\caption{\textbf{Cross-entropy after fine-tuning without Retroviridae.}
Top: tier-0 starting models. Bottom: tier-1 starting models. Dotted lines mark the same-size
unfiltered references. The human-infecting panels use 535 non-Retroviridae
sequences; other sequence strata are unchanged. }
\label{fig:retroviral-trajectories}
\end{figure}

\begin{figure}[p]
\centering
\includegraphics[width=\textwidth,height=0.84\textheight,keepaspectratio]{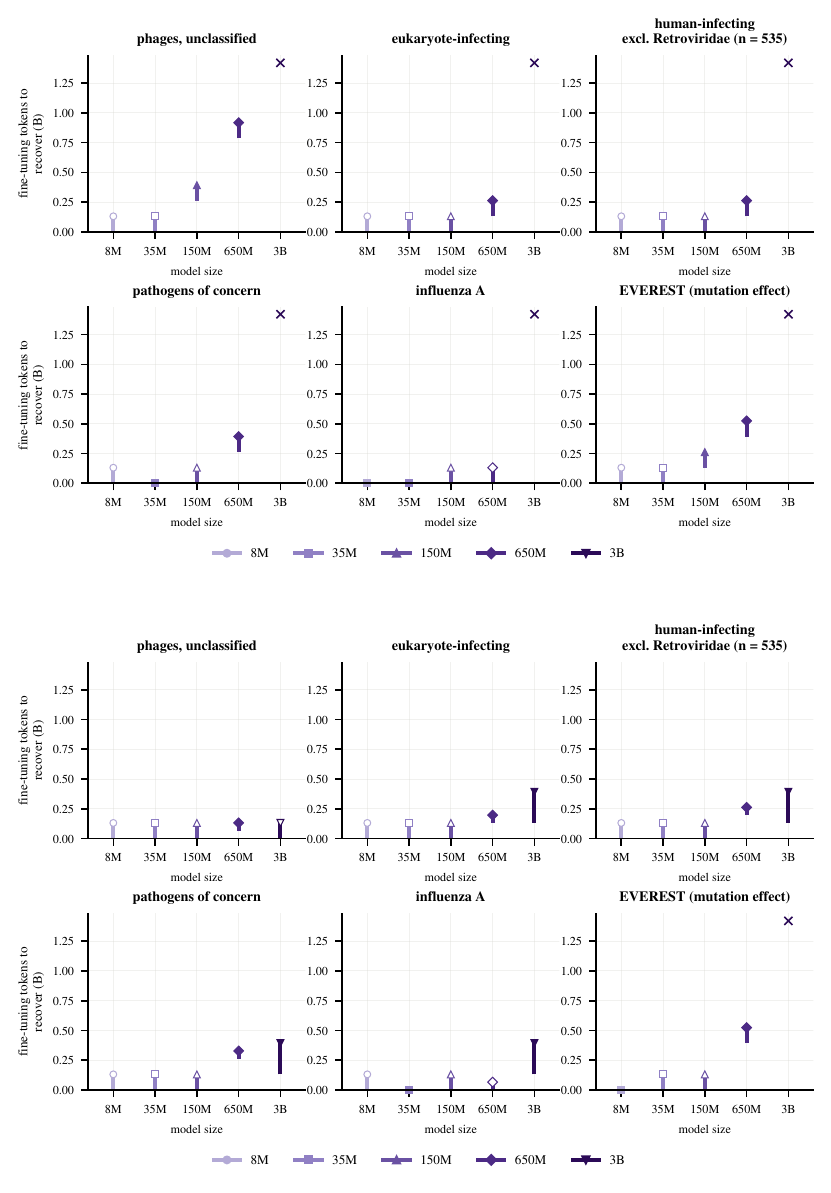}
\caption{\textbf{Tokens to recover performance without Retroviridae.}
Top: tier-0 starting models. Bottom: tier-1 starting models. Brackets span adjacent evaluations
around first attainment of the same-size unfiltered reference, without
interpolation. Open markers: first-checkpoint upper bounds; squares at zero:
initial parity; NR triangles: target not reached; crosses: not run. The human-infecting panels use 535 non-Retroviridae
sequences; EVEREST uses mean signed Spearman over all 45 assays.}
\label{fig:retroviral-recovery-cost}
\end{figure}

We test how well models fine-tuned without Retroviridae recover performance on held-out data excluding Retroviridae sequences (\Cref{fig:retroviral-appendix,fig:retroviral-trajectories,fig:retroviral-recovery-cost}, as described in \Cref{sec:app-retroviral-methods}).
Across 8M--650M parameters, models fine-tuned without Retroviridae from both tier-0 and tier-1
starting checkpoints recover mean EVEREST correlation and cross-entropy on the
human-infecting non-Retroviridae subset to the same-size unfiltered reference
within 2{,}000 fine-tuning steps. Mean EVEREST correlation reaches or exceeds the reference within the
first 500 steps at 8M and 35M, within 1{,}000 steps at 150M, and within 2{,}000
steps at 650M. Held-out cross-entropy on the human-infecting non-Retroviridae
stratum (535 sequences) follows the same course, i.e., recovery within 500 steps at 8M
and 35M, within 1{,}500 steps at 150M, and within 2{,}000 steps at 650M.
Recovery varies across other sequence strata; the tier-0 650M phage and
unclassified stratum recovers between 3{,}000 and 3{,}500 steps.

\clearpage

\subsection{Filtering effects over the course of pretraining}
\label{sec:app-pretrain-budget}

We score the frozen snapshots saved every 25{,}000 pretraining steps of the 650M unfiltered, tier-0 and tier-1 models, and the stock ESM-2 650M checkpoint, on the full validation split of our corpus (253{,}870 sequences). This split is held out from training for every filter. We compute the masked language-modelling loss with the training objective (\Cref{sec:app-pretraining}). Masks are seeded per sequence, so every checkpoint is scored on the same masked positions and differences between models are paired. We split the loss into \textcolor{forgetorange}{viral} sequences, the 7{,}668 validation sequences in a tier 1--5 held-out shell, and \textcolor{retainblue}{non-viral} sequences, the remainder. Sequences assigned to the viral set only by the tier-0 homology search are counted as non-viral. We fit $L(s) = E + A\,s^{-\alpha}$ to each curve and extrapolate to 500{,}000 steps. Extrapolations are indicative only, since the fit does not model the learning-rate schedule.

\begin{figure}[htbp]
\centering
\includegraphics[width=\textwidth]{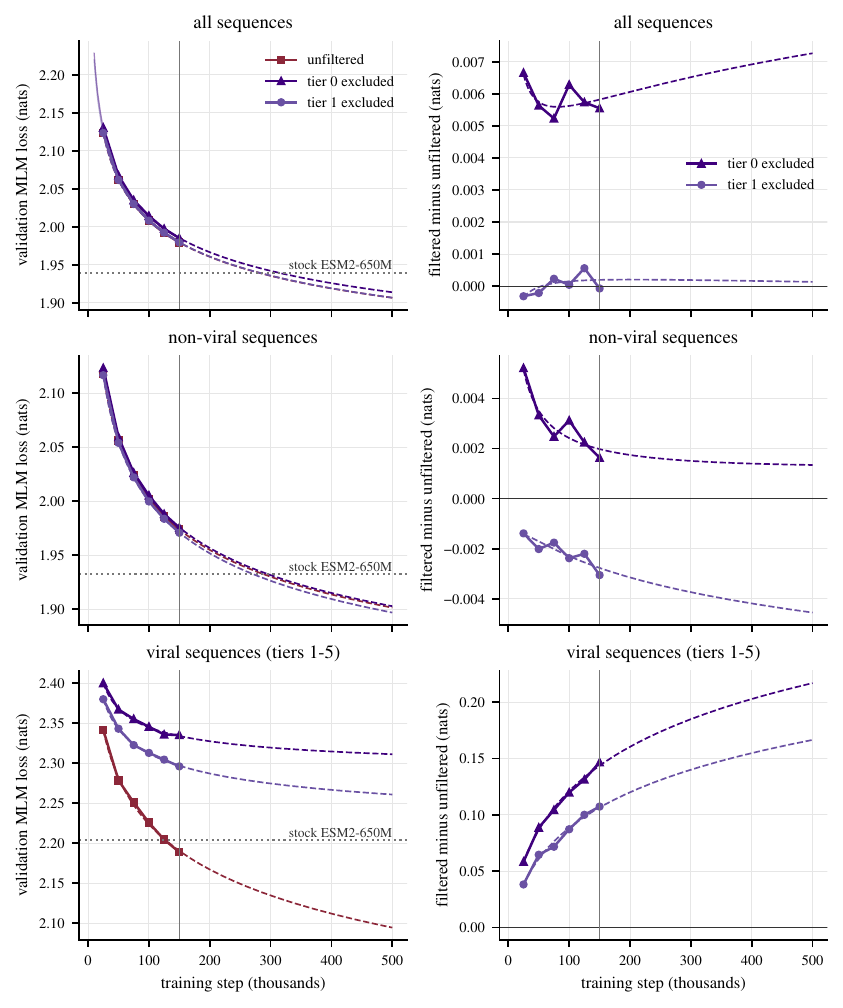}
\caption{\textbf{At 650M, the increase in \textcolor{forgetorange}{viral} validation loss from filtering grows over pretraining, while the \textcolor{retainblue}{non-viral} increase stays below 0.006 nats.} Masked language-modelling loss on the 253{,}870-sequence validation split, for snapshots every 25{,}000 steps. Rows: all, \textcolor{retainblue}{non-viral} and \textcolor{forgetorange}{viral} (tier 1--5 shells) sequences. Left: loss. Right: filtered minus unfiltered loss on the same masked positions. Solid lines connect rescored snapshots; dashed lines extrapolate a fit $L = E + A\,s^{-\alpha}$ to 500{,}000 steps. Faint lines (top left) show the validation loss logged during training. Dotted lines: stock ESM-2 650M (which may have trained on these sequences due to differences in how we filter for homology between train and test). The vertical line marks the end of our training at 150{,}000 steps. }
\label{fig:pretrain-budget-val-mlm}
\end{figure}

At 650M, the increase in \textcolor{forgetorange}{viral} loss from filtering grows at every snapshot (\Cref{fig:pretrain-budget-val-mlm}). Tier-0 minus unfiltered rises from $0.058$ nats at 25{,}000 steps to $0.147$ at 150{,}000 steps, and tier-1 minus unfiltered from $0.038$ to $0.107$. Over the last 50{,}000 steps the tier-0 model's viral loss falls by $0.010$ nats and the unfiltered model's by $0.037$. On \textcolor{retainblue}{non-viral} sequences, the tier-0 difference falls from $0.005$ to $0.002$ nats and the tier-1 difference stays between $-0.003$ and $-0.001$ nats. Stock ESM-2 650M, trained for 500{,}000 steps, has lower \textcolor{retainblue}{non-viral} loss than our unfiltered model at 150{,}000 steps ($1.932$ vs $1.974$ nats) and higher \textcolor{forgetorange}{viral} loss ($2.204$ vs $2.189$); it may have trained on these sequences due to differences in how we filter for homology between train and test.

\begin{figure}[htbp]
\centering
\includegraphics[width=\textwidth]{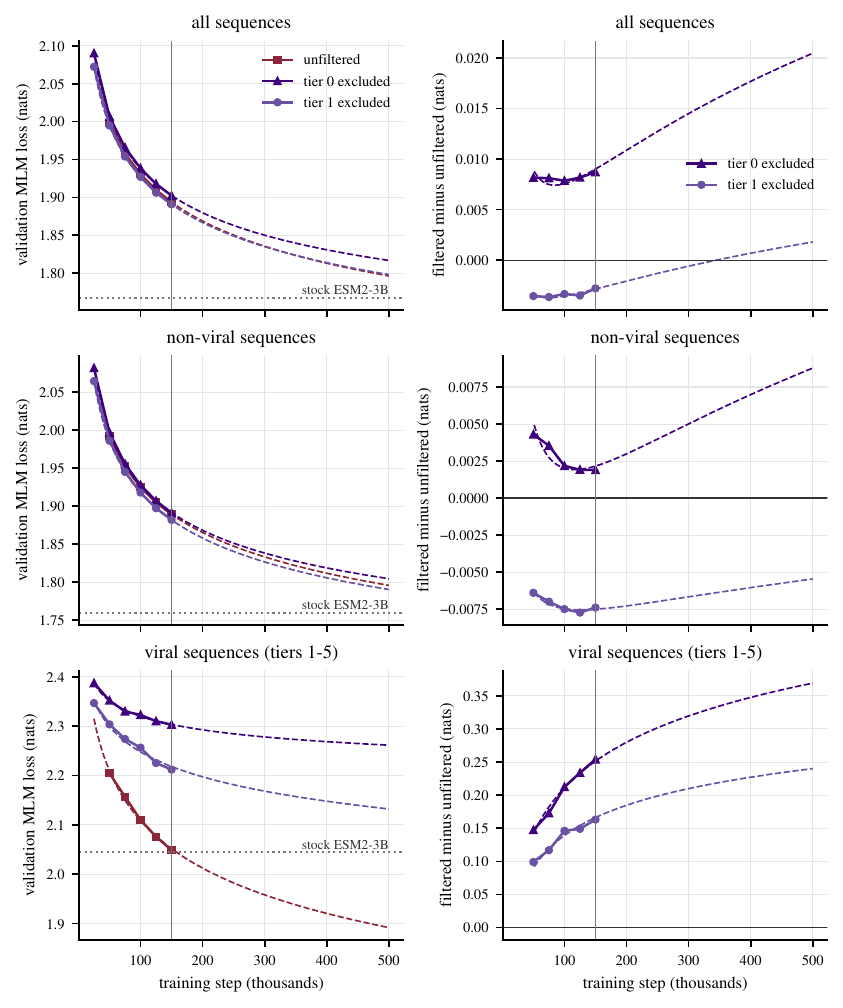}
\caption{\textbf{At 3B, the increase in \textcolor{forgetorange}{viral} validation loss from filtering also grows over pretraining.} Complementary analysis to \Cref{fig:pretrain-budget-val-mlm}, for the 3B models. Masked language-modelling loss on the 253{,}870-sequence validation split, for snapshots every 25{,}000 steps, starting from checkpoint at 50{,}000 steps. Rows: all, \textcolor{retainblue}{non-viral} and \textcolor{forgetorange}{viral} (tier 1--5 shells) sequences. Left: loss. Right: filtered minus unfiltered loss on the same masked positions. Solid lines connect rescored snapshots; dashed lines extrapolate a fit $L = E + A\,s^{-\alpha}$ to 500{,}000 steps. Faint lines (top left) show the validation loss logged during training. Dotted lines: stock ESM-2 3B (which may have trained on these sequences due to differences in how we filter for homology between train and test). The vertical line marks the end of our training at 150{,}000 steps.}
\label{fig:pretrain-budget-val-mlm-3b}
\end{figure}

\clearpage

\subsection{Fitness and stability prediction}
\label{sec:app-fitness-stability}

\begin{figure}[htbp]
\centering
\includegraphics[width=\textwidth]{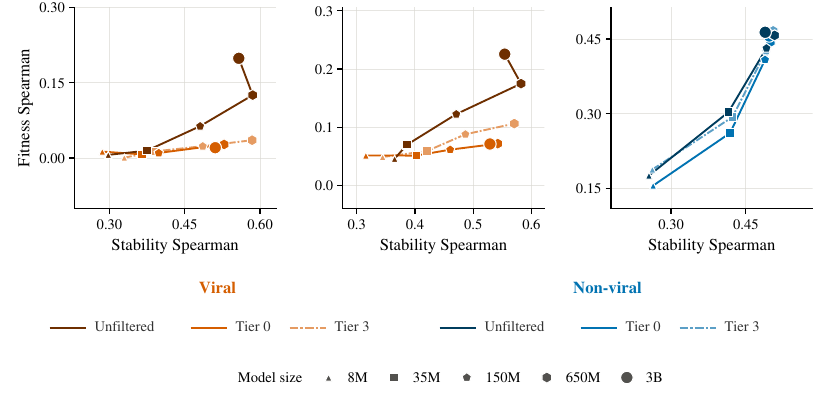}
\caption{\textbf{Data filtering degrades fitness prediction more than stability prediction.}
Left to right: ProteinGym viral, EVEREST viral and ProteinGym non-viral assays.
Each point shows the equally weighted mean Spearman correlation on stability assays (horizontal) and fitness assays (vertical) for one pretrained model (higher is better on both axes).
Orange: \textcolor{forgetorange}{viral} assays, blue \textcolor{retainblue}{non-viral} assays.
Fitness/stability assay counts are respectively 21/6, 31/5 and 56/60 in the three panels.
Fitness and stability use different assay sets; differences in their means may also reflect assay composition and difficulty.}
\label{fig:fitness-stability-planes}
\end{figure}

\begin{figure}[htbp]
\centering
\includegraphics[width=\textwidth]{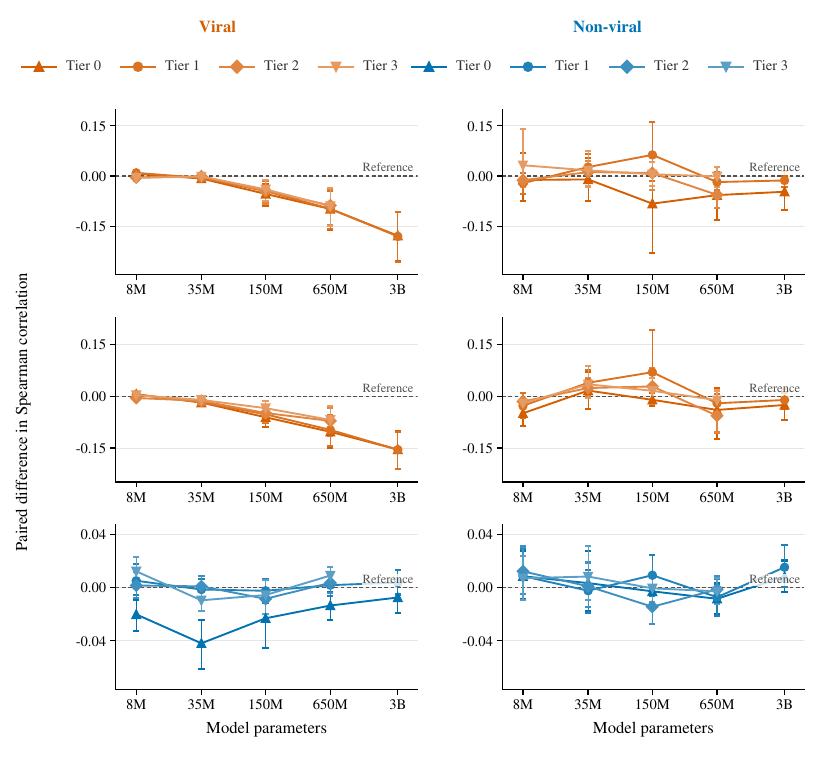}
\caption{\textbf{Filtering effects on fitness and stability prediction per tier.}
Rows show ProteinGym viral, EVEREST viral and ProteinGym non-viral assays; columns show fitness (left) and stability (right).
Points are equally weighted mean Spearman differences, filtered minus the same-size unfiltered reference, paired by assay.
Bars are pointwise 95\% percentile intervals from 20{,}000 paired-assay bootstrap resamples (seed~0).
Negative values indicate lower predictive performance; the dashed zero line denotes reference performance.
Orange: \textcolor{forgetorange}{viral} assays, blue \textcolor{retainblue}{non-viral} assays.
Fitness/stability counts are 21/6, 31/5 and 56/60 by row, and the two columns share limits within each row.}
\label{fig:fitness-stability-filtering}
\end{figure}

\clearpage
\subsection{Assay-level recovery after viral fine-tuning}
\label{sec:app-assay-recovery}

\begin{figure}[htbp]
\centering
\includegraphics[width=\textwidth]{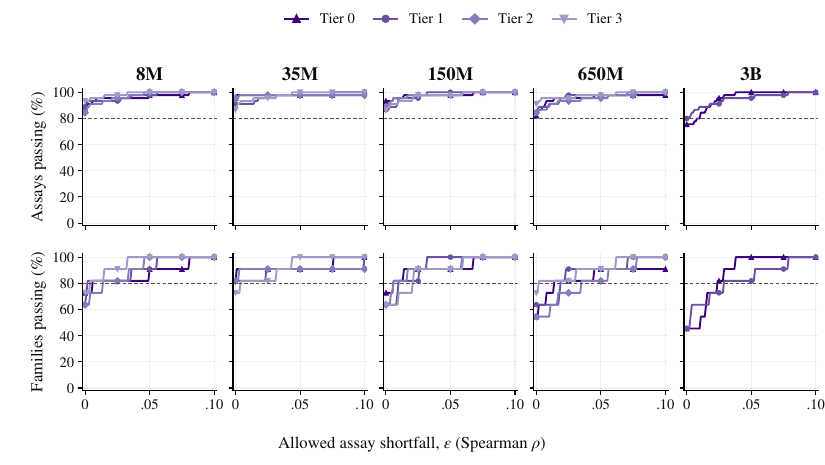}
\caption{\textbf{Sensitivity to the allowed assay-level shortfall.}
Columns show model size; colours and marker shapes identify the starting filter tier.
At the fixed step-5{,}000 checkpoint after ordinary viral fine-tuning, an assay passes when its signed Spearman correlation is at least the same-size unfiltered reference model's correlation minus $\epsilon$.
The top row shows the percentage of the 45 assays passing. The bottom row shows the percentage of the 11 families for which every assay within the family passes.
Assays have equal weight in the top row and families have equal weight in the bottom row.
Dashed horizontal lines mark 80\%, corresponding to at least 9 of 11 families in the bottom row.
The allowed shortfall $\epsilon$ is an absolute Spearman difference from 0 to 0.10; at $\epsilon=0$, passing requires parity or better.}
\label{fig:assay-recovery-tolerance}
\end{figure}

\begin{figure}[htbp]
\centering
\includegraphics[width=\textwidth]{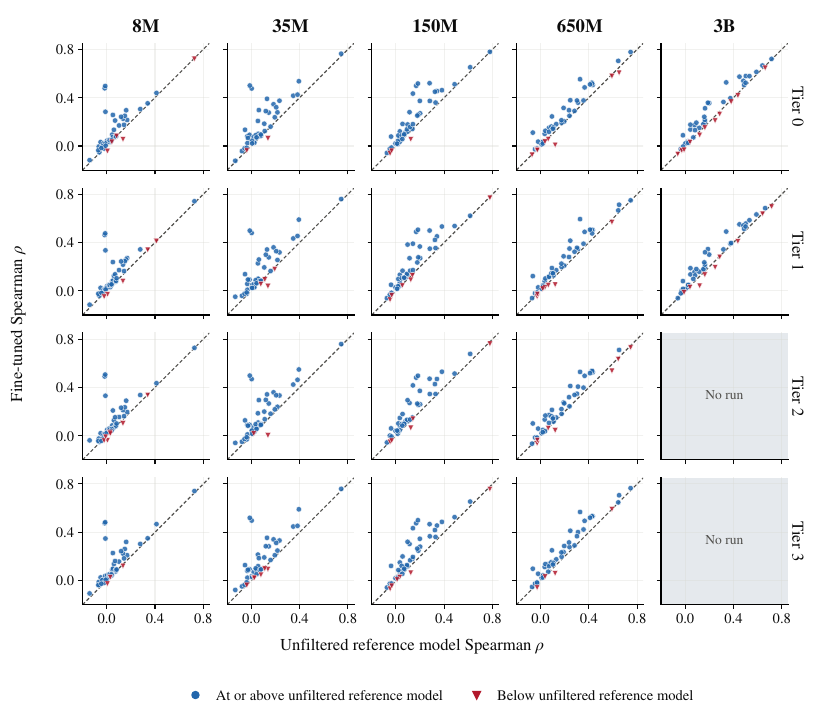}
\caption{\textbf{Unfiltered versus fine-tuned performance by EVEREST assay.}
Columns show model size and rows show the starting filter tier.
The horizontal axis is the same-size unfiltered reference model's signed Spearman correlation;
the vertical axis is the correlation after fine-tuning to recover degraded performance (\Cref{fig:recovery-crossing}) at step 5{,}000 (1.311 billion training tokens).
The dashed diagonal marks parity. Blue circles are at or above the unfiltered reference model, and red downward triangles are below it.}
\label{fig:assay-recovery-scatter}
\end{figure}

\begin{figure}[htbp]
\centering
\includegraphics[width=\textwidth]{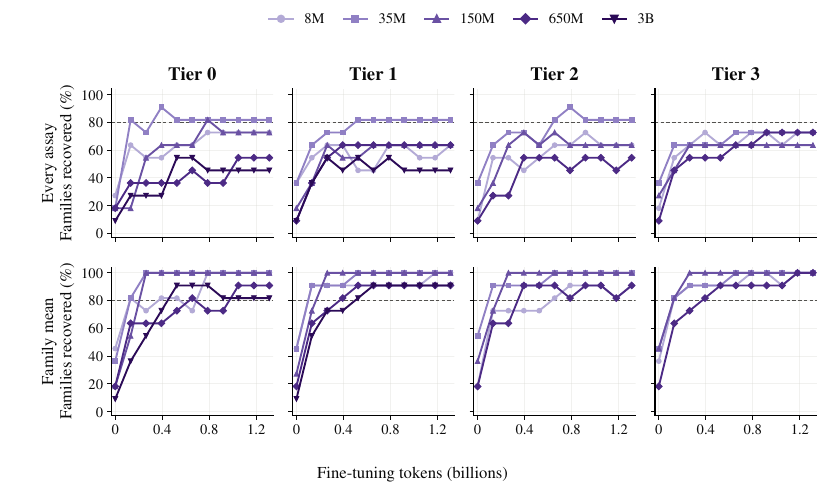}
\caption{\textbf{Viral-family recovery during fine-tuning on viral sequences.}
Columns show the starting filter tier; colours and markers identify model size.
At each checkpoint, the vertical axis gives the percentage of the 11 evaluated EVEREST families whose mutation-effect prediction performance matches or exceeds that of the same-size unfiltered reference model.
The top row requires every assay in a family to match or exceed its own unfiltered reference model score.
The bottom row requires the mean signed Spearman correlation across that family's assays to match or exceed the corresponding mean for the unfiltered reference model.
Families have equal weight in both rows and we weight assays equally for each family mean.
Dashed lines mark 80\%, requiring at least 9 of 11 families at the same checkpoint.
Markers show the starting models and checkpoints every 500 steps (131{,}072{,}000 tokens), through step 5{,}000 (1{,}310{,}720{,}000 tokens).
Families already meeting the criterion at step zero are included. We recalculate coverage at each checkpoint and it may subsequently fall.}
\label{fig:assay-recovery-trajectories}
\end{figure}

\begin{figure}[p]
\centering
\includegraphics[width=\textwidth]{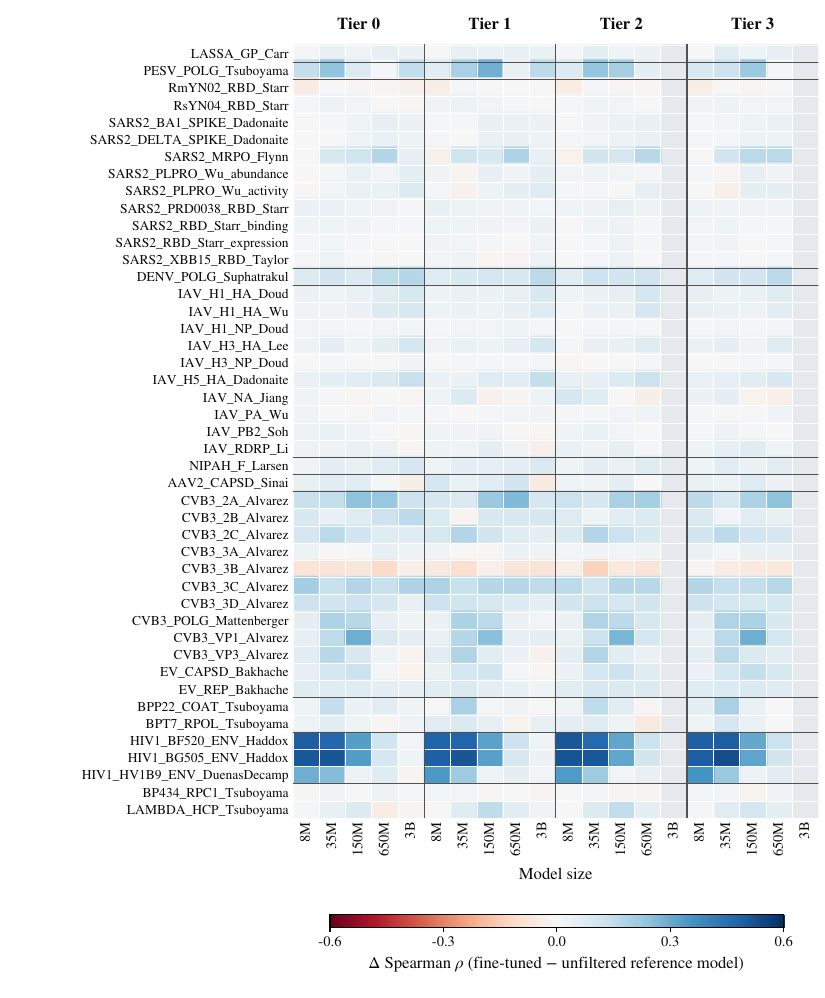}
\caption{\textbf{Per-assay performance after fine-tuning relative to the unfiltered reference model.}
Each cell is the final signed Spearman correlation minus the same-size unfiltered reference model's correlation.
Blue denotes higher performance, red lower performance and white parity. Grey denotes no evaluated run.
The symmetric colour scale spans $-0.6$ to $+0.6$.
Rows are the 45 EVEREST assays, grouped by viral family and ordered by identifier within each family.
Columns group model sizes within starting filter tiers.
All values are from step 5{,}000 of viral fine-tuning (1.311 billion training tokens).}
\label{fig:assay-recovery-differences}
\end{figure}

In \Cref{fig:recovery-crossing}, we examine the cost of fine-tuning to recover mean performance on the held-out viral shells and across EVEREST assays. Here we explore recovery in mutation effect prediction within individual viral families and EVEREST assays. Recovery is heterogeneous: family means can recover while individual assays remain below the same-size unfiltered reference model (\Cref{fig:assay-recovery-trajectories}). At step 5{,}000, each evaluated model/tier combination has at most two of the 45 assays more than 0.05 Spearman below its unfiltered reference (\Cref{fig:assay-recovery-tolerance,fig:assay-recovery-scatter}). The largest residual shortfalls are concentrated in a small subset of assays, with a maximum shortfall of 0.136 Spearman (\Cref{fig:assay-recovery-differences}).

\clearpage

\clearpage

\clearpage

\clearpage

\subsection{Supervised fine-tuning of the filtered, unfiltered and recovered 3B models}
\label{sec:app-partb}

We compare three 3B checkpoints under supervised fine-tuning on ProteinGym
labels. These are the unfiltered model, the tier-0 filtered model, and the
tier-0 model after 5{,}000 steps of continued pretraining on the withheld
\textcolor{forgetorange}{viral} corpus (``recovered'',
\Cref{sec:app-fine-tuning}). We apply LoRA
\citep{hu2021loralowrankadaptationlarge} to the query and value projections of
every layer, keeping the base weights frozen. The adapter output weights are
initialised to zero, so step 0 reproduces the base model.
\Cref{tab:sft-settings} lists the shared training settings.

\begin{table}[H]
\centering
\caption{\textbf{Supervised fine-tuning settings shared across checkpoints and regimes.}}
\label{tab:sft-settings}
\small
\renewcommand{\arraystretch}{1.12}
\begin{tabular}{ll}
\toprule
Setting & Value \\
\midrule
LoRA rank & 16 \\
LoRA $\alpha$ & 32 \\
Adapter dropout & 0.05 \\
Trainable parameters & 5{,}898{,}240 \\
Optimiser & AdamW \\
Learning rate & $10^{-4}$ \\
Fine-tuning steps & 3{,}000 \\
Seeds & 3 \\
Training variant cap & 2{,}000 per assay \\
Training window & 512 residues, centred on the mutation \\
Evaluation interval & 500 steps \\
\bottomrule
\end{tabular}
\end{table}

We train on single substitutions only, sampling assays uniformly and then
pairs of variants with unequal DMS labels within each assay. For a pair $a,b$,
we minimise the pairwise logistic ranking loss
\[
\mathcal{L}_{a,b}
= \mathrm{softplus}\!\left(-\mathrm{sign}(y_a-y_b)(s_a-s_b)\right),
\]
where $y_a$ and $y_b$ are the DMS labels and $s$ is the masked-marginal score
of \Cref{sec:app-mutation-effect}. We add no assay-specific head or
calibration, so evaluation uses the same masked-marginal scorer as the rest
of the paper.

We split assays by whole protein group. Groups are connected components
linked by shared UniProt identifiers, identical wild-type sequences, or
MMseqs2 alignments with identity $\geq 0.3$ and coverage $\geq 0.8$ of at
least one sequence. The whole-protein containment gate of
\Cref{sec:app-fine-tuning}, with identity and coverage $\geq 0.8$, finds no
cross-split pair. We assign 40\% of each viral family's groups to training.
\Cref{tab:sft-splits} gives the resulting split sizes.

\begin{table}[H]
\centering
\caption{\textbf{ProteinGym assay splits.}
Entries show assays (protein groups). There is no viral validation set.}
\label{tab:sft-splits}
\small
\renewcommand{\arraystretch}{1.12}
\begin{tabular}{@{}lccc@{}}
\toprule
Domain & Training & Validation & Test \\
\midrule
\textcolor{forgetorange}{Viral} & 13 (8) & --- & 18 (14) \\
\textcolor{retainblue}{Non-viral} & 122 (93) & 28 (20) & 36 (20) \\
\bottomrule
\end{tabular}

\end{table}

The two regimes differ only in their training pool. The \emph{mixed} regime
uses all 135 training assays, while the \emph{benign-only} regime uses the
122 \textcolor{retainblue}{non-viral} assays and excludes the 13
\textcolor{forgetorange}{viral} training assays. Both regimes use the same
split, three seeds and held-out test sets. Within each regime, all three
checkpoints use the same fixed training pool.

We evaluate all valid substitutions, including multi-site mutations, at each
evaluation step. Confidence intervals use 10{,}000-resample bootstraps over
crossed protein-group and seed axes, preserving the pairing between models.
These intervals are conditional on the fixed checkpoints. For paired tests,
we first average seeds within each assay, then apply two-sided sign-flip
permutation tests over protein groups. We apply a Holm correction within
each family of tests.

\subsection{Does the recovered model behave like the unfiltered model?}
\label{sec:app-partb-recovered}

\begin{figure}[H]
\centering
\includegraphics[width=\textwidth]{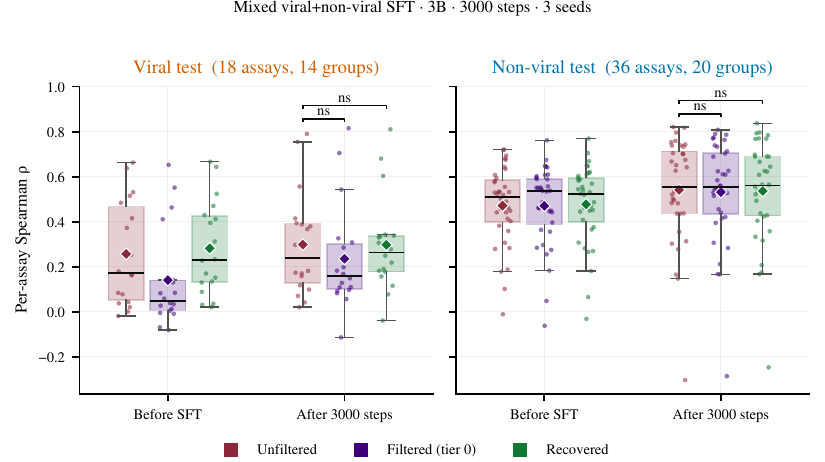}
\caption{\textbf{Per-assay Spearman $\rho$ for the
unfiltered, filtered and recovered 3B models before and after mixed fine-tuning,
on held-out \textcolor{forgetorange}{viral} and \textcolor{retainblue}{non-viral}
assays.} Boxes average the three seeds per assay. Diamonds mark assay means.
Brackets at ``after'' test the filtered-minus-unfiltered and
recovered-minus-unfiltered contrasts.}
\label{fig:sft-mixed-box}
\end{figure}

\begin{figure}[H]
\centering
\includegraphics[width=\textwidth]{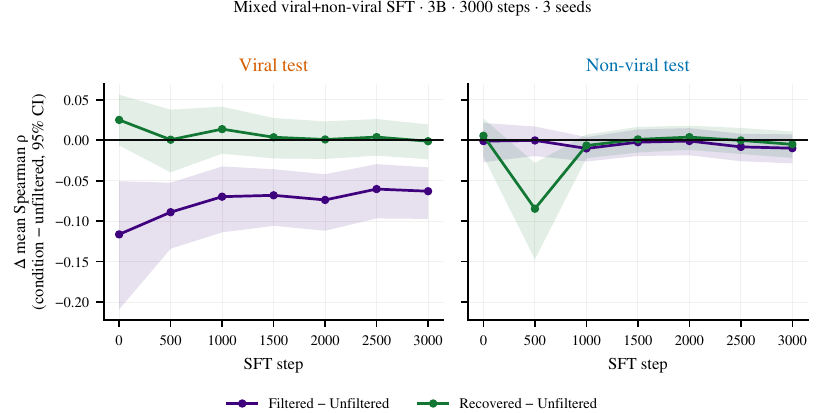}
\caption{\textbf{Filtered-minus-unfiltered and
recovered-minus-unfiltered Spearman $\rho$ across mixed fine-tuning steps.} Paired
per-assay contrasts with 95\% bootstrap bands, one panel per domain. Zero marks
parity with the unfiltered model under the same fine-tuning.}
\label{fig:sft-mixed-penalty-lines}
\end{figure}

\begin{figure}[H]
\centering
\includegraphics[width=0.8\textwidth]{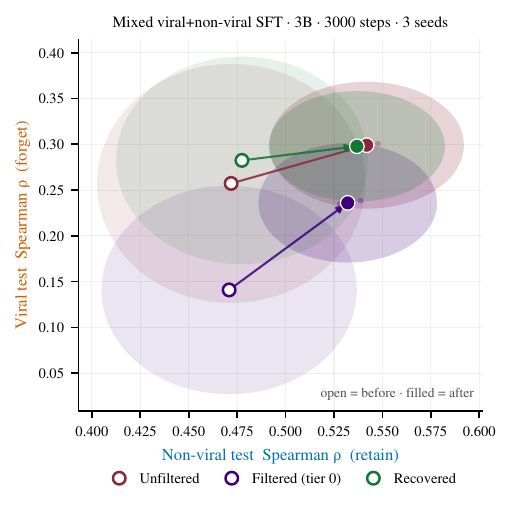}
\caption{\textbf{Joint \textcolor{forgetorange}{viral}
and \textcolor{retainblue}{non-viral} Spearman $\rho$ for each condition, before vs
after mixed fine-tuning.} Each arrow runs from the pre- to the post-fine-tuning
mean for one condition, with 95\% confidence ellipses at each endpoint.}
\label{fig:sft-mixed-planes}
\end{figure}

Under mixed fine-tuning we compare the
unfiltered, filtered and recovered 3B checkpoints on the same held-out assays
(\Cref{fig:sft-mixed-box,fig:sft-mixed-planes}). Before fine-tuning, the filtered
model is 0.116 Spearman below the unfiltered model on the
\textcolor{forgetorange}{viral} assays ($p=0.009$ after Holm correction), while the
recovered model is 0.025 above it ($p=1.0$). After 3{,}000 steps the filtered model
remains 0.063 below the unfiltered model (Holm-adjusted $p=0.12$) and the recovered
model is within 0.001 of it ($p=1.0$). On the \textcolor{retainblue}{non-viral}
assays no contrast exceeds 0.010 Spearman at either stage, and none is significant.
The same ordering holds across the trajectory
(\Cref{fig:sft-mixed-penalty-lines}): the recovered-minus-unfiltered band covers
zero at every step, and the filtered-minus-unfiltered gap narrows but stays
negative on the \textcolor{forgetorange}{viral} assays. Therefore, the recovered model behaves very similarly to the unfiltered model. The
exception is a transient drop at the first evaluation, at step 500, in one of the
three seeds: its recovered model scores 0.244 mean Spearman on the
\textcolor{retainblue}{non-viral} assays against 0.509 and 0.513 for the other two
seeds, and 0.523 by step 1{,}000. Averaged over seeds this leaves the
recovered-minus-unfiltered contrast 0.085 below zero at step 500 and back to
$-0.007$ at step 1{,}000 (\Cref{fig:sft-mixed-penalty-lines}). The same seed dips
only 0.038 on the \textcolor{forgetorange}{viral} assays.

\begin{figure}[H]
\centering
\includegraphics[width=\textwidth]{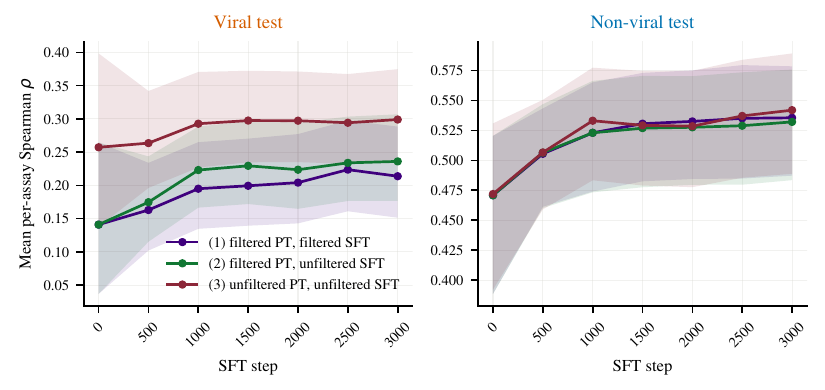}
\caption{\textbf{Mean per-assay Spearman $\rho$ across
fine-tuning steps for three pretraining\,$\times$\,fine-tuning combinations}. Compare: 
filtered pretraining with \textcolor{retainblue}{non-viral} fine-tuning, filtered
pretraining with mixed fine-tuning, and unfiltered pretraining with mixed
fine-tuning. The full held-out test set is scored every 500 steps, with 95\%
bootstrap bands. Panels split the \textcolor{forgetorange}{viral} and
\textcolor{retainblue}{non-viral} domains.}
\label{fig:sft-threeway}
\end{figure}

Holding the fine-tuning data fixed and varying the pretraining corpus
(\Cref{fig:sft-threeway}), the unfiltered model ends 0.063 Spearman
(95\% CI 0.033--0.097) above the filtered model on the held-out
\textcolor{forgetorange}{viral} assays and 0.010 (95\% CI $-0.007$--0.029) above it
on the \textcolor{retainblue}{non-viral} assays. Mixed fine-tuning of the filtered
model therefore closes part of the \textcolor{forgetorange}{viral} gap left by
filtering, whereas fine-tuning after recovery on the withheld corpus leaves no
measurable gap. However, we note that there is also improvement in performance when training on the filtered data, which we explore further in \Cref{sec:app-partb-benign}.

\subsection{Does benign fine-tuning recover performance?}
\label{sec:app-partb-benign}

\begin{figure}[H]
\centering
\includegraphics[width=\textwidth]{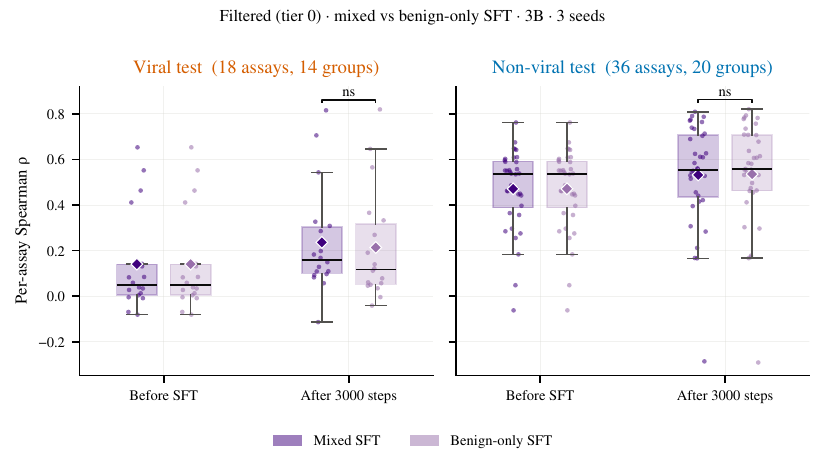}
\caption{\textbf{Filtered-model per-assay Spearman
$\rho$ before and after fine-tuning, under the benign-only and mixed regimes.} Each
cluster places the two regimes side by side on the held-out
\textcolor{forgetorange}{viral} and \textcolor{retainblue}{non-viral} assays. Boxes
average the three seeds per assay. Diamonds mark assay means.}
\label{fig:sft-benign-box}
\end{figure}

 Mixed fine-tuning trains on both \textcolor{forgetorange}{viral} and \textcolor{retainblue}{non-viral} assays. We next ask how far the filtered model
recovers on \textcolor{forgetorange}{viral} assays when the fine-tuning data contains no \textcolor{forgetorange}{viral}
assay at all. Benign-only fine-tuning raises the
filtered model's mean \textcolor{forgetorange}{viral} Spearman from 0.141 to 0.214
over 3{,}000 steps. Mixed fine-tuning reaches 0.236
(\Cref{fig:sft-benign-box}).

\begin{figure}[H]
\centering
\includegraphics[width=\textwidth]{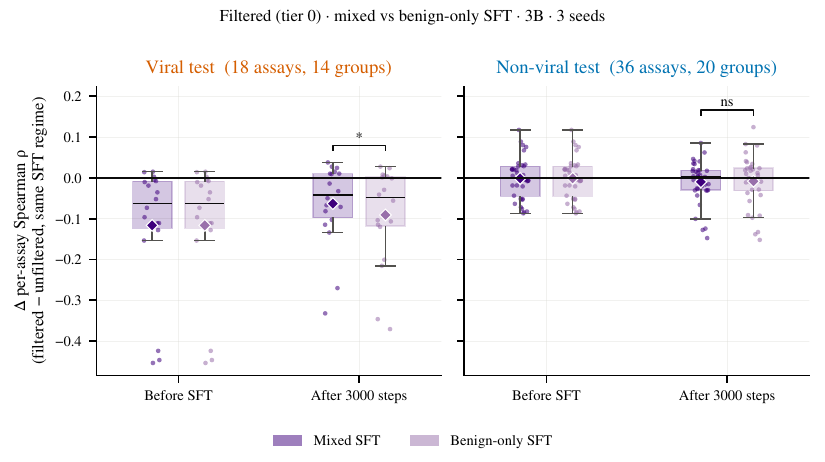}
\caption{\textbf{Per-assay filtered-minus-unfiltered
Spearman $\rho$ before and after fine-tuning, for the benign-only and mixed
regimes.} Negative values indicate that filtering lowers performance. Each
difference subtracts the unfiltered model under the same regime, step, seed and
assay.}
\label{fig:sft-benign-penalty-box}
\end{figure}

\begin{figure}[H]
\centering
\includegraphics[width=\textwidth]{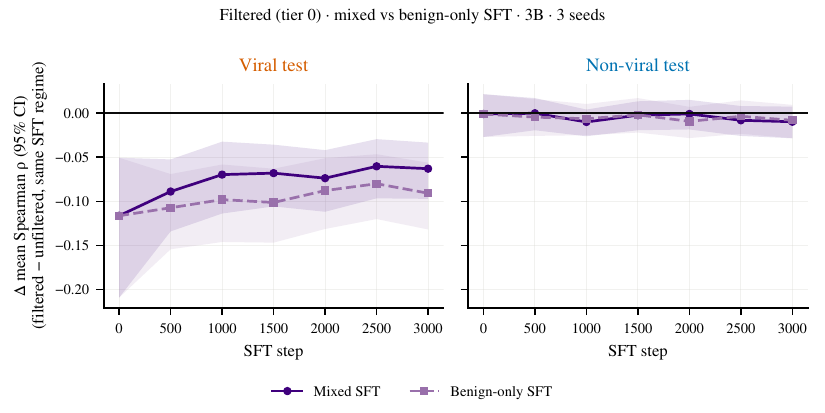}
\caption{\textbf{Filtered model does not catch up to unfiltered model under the same fine-tuning regime, even with viral data.} Filtered-minus-unfiltered Spearman
$\rho$ across fine-tuning steps for the benign-only and mixed regimes. Paired 95\%
bootstrap bands, one panel per domain.}
\label{fig:sft-benign-penalty-lines}
\end{figure}

Against the unfiltered model trained under the same regime, the
\textcolor{forgetorange}{viral} gap narrows from $-0.116$ to $-0.091$ under
benign-only fine-tuning, and remains significant (Holm-adjusted $p=0.03$). Under
mixed fine-tuning it narrows to $-0.063$, which is not significant after
correction ($p=0.12$). The two regimes differ by 0.028 Spearman (Holm-adjusted
$p=0.045$) (\Cref{fig:sft-benign-penalty-box,fig:sft-benign-penalty-lines}).

\begin{figure}[H]
\centering
\includegraphics[width=\textwidth]{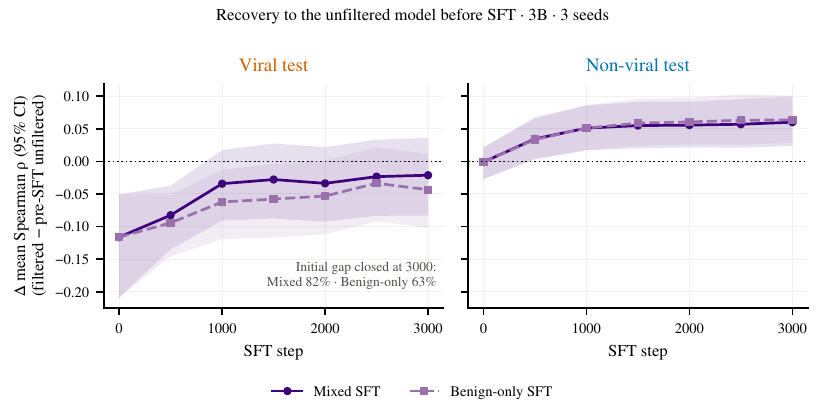}
\caption{\textbf{Filtered model recovers performance relative to unfiltered model at the end of pretraining, even without viral data.} Gap from the filtered model to the
fixed pre-fine-tuning unfiltered reference across fine-tuning steps. One panel per
domain. Zero marks the unfiltered model's pre-fine-tuning performance, which is
shared by both regimes. Evaluations every 500 steps, with 95\% bootstrap bands on
the paired per-assay gap.}
\label{fig:sft-benign-recovery}
\end{figure}

Against the fixed pre-fine-tuning unfiltered model
(\Cref{fig:sft-benign-recovery}), benign-only fine-tuning closes 63\% of the
initial \textcolor{forgetorange}{viral} gap and mixed fine-tuning 82\%. Neither
reaches the reference within 3{,}000 steps.

On \textcolor{retainblue}{non-viral} assays both regimes pass the reference by the
first 500-step evaluation, ending 0.064 (benign-only) and 0.060 (mixed) Spearman
above it. Their \textcolor{retainblue}{non-viral} gaps to the unfiltered model
differ by 0.002 ($p=1.0$).

Benign fine-tuning therefore recovers some of the
\textcolor{forgetorange}{viral} performance that filtering removed, though less
than fine-tuning that includes \textcolor{forgetorange}{viral} assays. The
trajectories in \Cref{fig:sft-threeway} separate in the same order.

\subsection{Transfer to EVEREST}
\label{sec:app-partb-everest}

The fine-tuning above trains and tests on
ProteinGym, so the held-out \textcolor{forgetorange}{viral} assays share a
benchmark, and in places a protein family, with the fine-tuning data. We therefore
score the same adapters on EVEREST, which is entirely
\textcolor{forgetorange}{viral}. Each
EVEREST wild type is searched against the wild types of the 13
\textcolor{forgetorange}{viral} ProteinGym assays in the mixed fine-tuning pool
with the homology filter used to build the ProteinGym split (MMseqs2 identity $\geq 0.3$ at
$\geq 0.8$ coverage of either sequence, plus exact matches), which flags 16 of the
45 EVEREST assays as homologous to that pool. We further exclude every assay shared with ProteinGym, leaving 16 assays in 5
homology groups that neither appear in ProteinGym nor are homologous to anything
fine-tuned on. Every 500-step adapter from both regimes and all three seeds is
scored on these assays with the unchanged masked-marginal scorer, and reported as
the unweighted mean of per-assay signed Spearman correlations with the bootstrap
and sign-flip procedures used above. 

\begin{figure}[H]
\centering
\includegraphics[width=\textwidth]{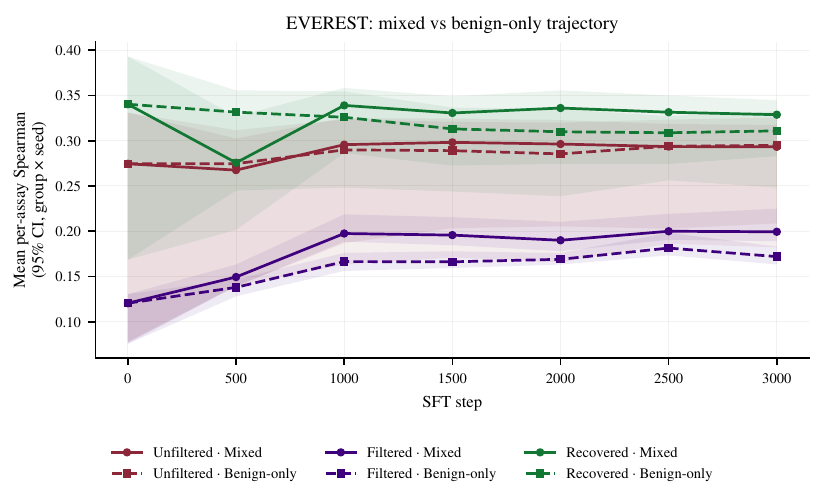}
\caption{\textbf{Mean per-assay Spearman $\rho$ on
EVEREST across fine-tuning steps, for the unfiltered, filtered and recovered 3B
models under both fine-tuning regimes.} Solid lines are mixed fine-tuning, dashed
lines benign-only. The 16 EVEREST assays shown are absent from ProteinGym and
non-homologous to the fine-tuning pool. Bands are 95\% bootstrap intervals over
the five homology groups and three seeds.}
\label{fig:everest-novel-lines}
\end{figure}

\begin{figure}[H]
\centering
\includegraphics[width=\textwidth]{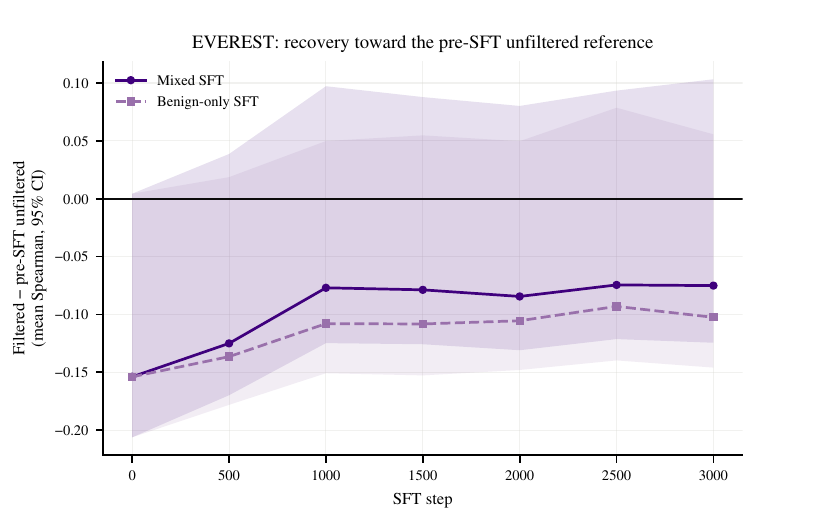}
\caption{\textbf{Gap from the filtered model to the
pre-fine-tuning unfiltered reference on EVEREST across fine-tuning steps.} Zero
marks the unfiltered model's EVEREST performance before fine-tuning, shared by
both regimes. The 16 assays and bootstrap are as in 
\Cref{fig:everest-novel-lines}.}
\label{fig:everest-novel-recovery}
\end{figure}

The ordering carries over to EVEREST
(\Cref{fig:everest-novel-lines}). Before fine-tuning, the filtered model is 0.154
Spearman below the unfiltered model and the recovered model is 0.066 above it. At
3{,}000 steps of mixed fine-tuning the filtered model is 0.094 below and the
recovered model 0.035 above. Under benign-only fine-tuning the filtered model is
0.123 below and the recovered model 0.016 above. Measured against the unfiltered
model before fine-tuning (\Cref{fig:everest-novel-recovery}), the filtered model
closes 51\% of its initial gap under mixed fine-tuning and 33\% under benign-only
fine-tuning. Fine-tuning on ProteinGym labels therefore
transfers to \textcolor{forgetorange}{viral} assays that are neither in the
fine-tuning data nor homologous to it, and recovers part but not all of what
filtering removed, as on the ProteinGym held-out assays. The recovered model
stays at or above the unfiltered model throughout.

\end{document}